\documentclass[11pt]{article}
\usepackage[final]{acl}
\usepackage{times}
\usepackage{latexsym}
\usepackage[T1]{fontenc}
\usepackage[utf8]{inputenc}
\usepackage{microtype}
\usepackage{inconsolata}
\usepackage{graphicx}
\usepackage{booktabs}
\usepackage{multirow}
\usepackage{wrapfig}
\usepackage{multicol}
\usepackage{amsmath}
\usepackage{url}
\usepackage{array}
\usepackage{enumitem}
\usepackage{array}
\usepackage{tabularx}
\usepackage[table,xcdraw]{xcolor}
\usepackage{fontspec}
\usepackage{polyglossia}
\usepackage{pgfplots}
\usetikzlibrary{shapes.geometric}
\pgfplotsset{compat=1.18}
\setmainlanguage{english}
\setotherlanguage{bengali}
\usepackage{fontspec}
\usepackage{polyglossia}
\usepackage{tikz}
\usepgfplotslibrary{groupplots}
\usetikzlibrary{calc}
\setmainlanguage{english}
\setotherlanguage{bengali}
\newfontfamily\bengalifont[Script=Bengali]{NotoSerifBengali-Regular.ttf}
\usepackage{placeins}
\definecolor{promptblue}{HTML}{D6EAF8}
\definecolor{promptpink}{HTML}{FADBD8} 
\definecolor{promptgreen}{HTML}{FEF9E7}  
\definecolor{promptyellow}{HTML}{E8DAEF}
\definecolor{cellbest}{HTML}{C8E6C9}  
\definecolor{cellworst}{HTML}{FFCDD2}

\title{BenHalluEval: A Multi-Task Hallucination Evaluation Framework for Large Language Models on Bengali}

\author{
  \textbf{Shefayat E Shams Adib}$^{\dagger}$\thanks{Equal contribution.},
  \textbf{Ahmed Alfey Sani}$^{\dagger}$\footnotemark[1],
  \textbf{Ekramul Alam Esham}$^{\dagger}$\footnotemark[1] \\
  \textbf{Ajwad Abrar}$^{\dagger}$,
  \textbf{Ishmam Tashdeed}$^{\dagger}$,
  \textbf{Md Taukir Azam Chowdhury}$^{\ddagger}$ \\[4pt]
  $^{\dagger}$Department of Computer Science and Engineering,
  Islamic University of Technology \\
  $^{\ddagger}$Department of Computer Science and Engineering,
  University of California \\[2pt]
  {\small \{shefayatadib, ahmedalfey, ekramulalam, ajwadabrar,
  ishmamtashdeed\}@iut-dhaka.edu, mchow068@ucr.edu}
}

\begin{document}
\maketitle

\begin{abstract}
Despite Bengali being the sixth most spoken language in the world,
no prior work has systematically evaluated hallucination in large
language models (LLMs) for Bengali. We introduce \textbf{BenHalluEval},
a hallucination evaluation framework covering four tasks: Question
Answering (QA), Bangla-English Code-Mixed QA, Summarization, and
Mathematical Reasoning. We construct 12,000 hallucinated candidates
using GPT-5.4 across twelve hallucination types and evaluate nine
LLMs under a dual-track protocol that independently measures
false-positive rate on ground-truth instances (Track~A) and
hallucination detection rate on hallucinated candidates (Track~B).
Three native Bengali annotators validate the benchmark at
almost-perfect agreement (Fleiss' $\kappa = 0.911$--$0.926$). To
jointly penalise both failure modes, we propose \textbf{BenHalluScore},
a dual-track metric equivalent to balanced error rate. BenHalluScore
ranges from 7.72\% to 55.42\%, with rankings task-conditional rather
than scale-ordered. Chain-of-thought prompting lowers BenHalluScore
in 14 of 32 evaluable combinations and raises it in 18, with gains
on Summarization and Mathematical Reasoning but degradation on both
QA variants. BenHalluEval establishes the first dedicated
hallucination benchmark for Bengali and highlights the inadequacy of
single-track and prompting-only approaches for low-resource settings.
The dataset and code are available at
\url{https://anonymous.4open.science/r/BanglaHalluEval-EB77}.
\end{abstract}

\section{Introduction}
\label{sec:intro}

\begin{figure}[t]
  \centering
  \includegraphics[width=\columnwidth]{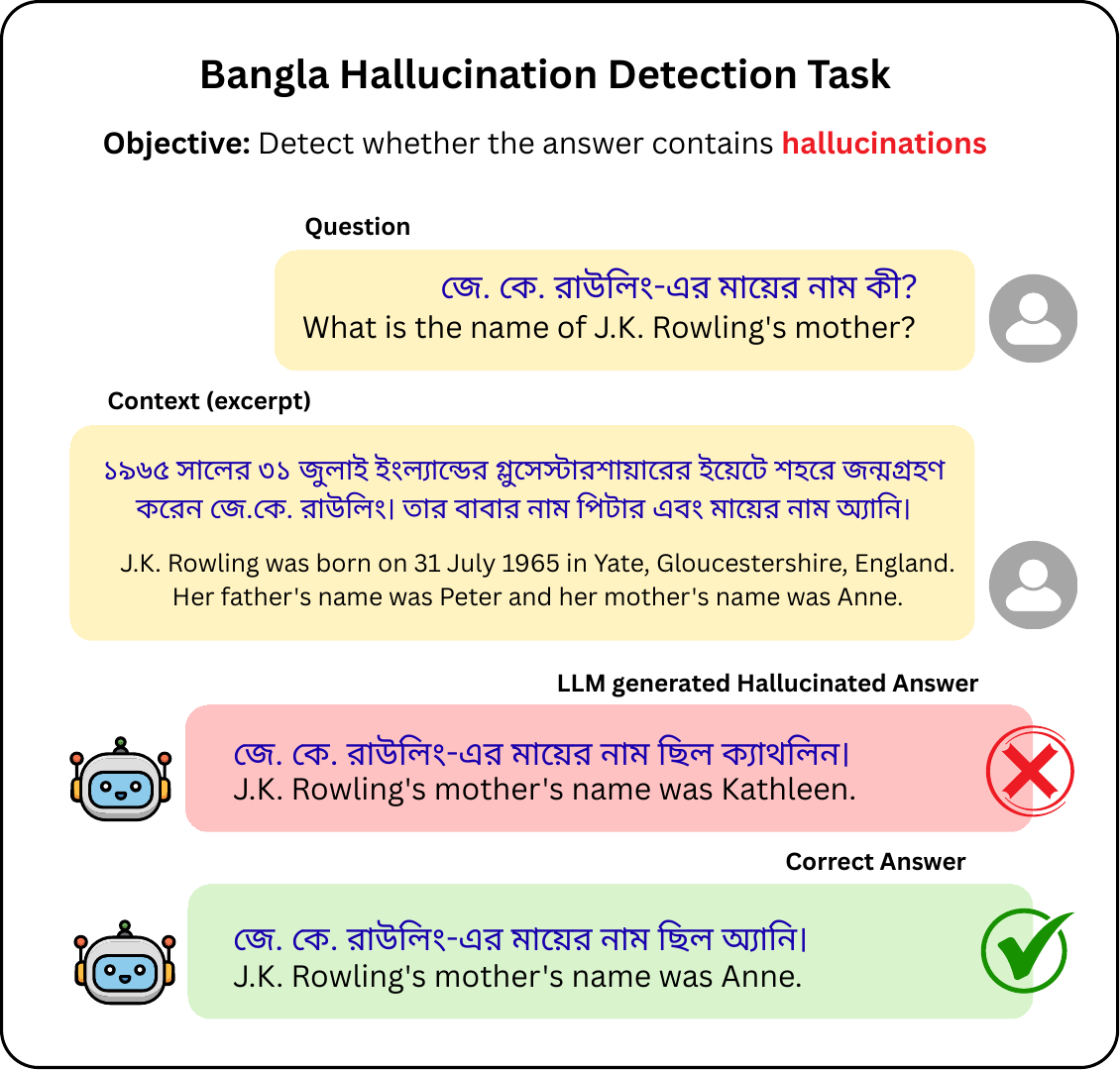}
  \caption{An example of hallucination detection 
  in BenHalluEval.}
  \label{fig:intro}
\end{figure}

Large language models (LLMs) generate outputs that contradict established facts, fabricate details, or conflict with the provided context, known as hallucination \cite{huang2023survey,li2023halueval}. A medical chatbot that invents drug dosages, a legal assistant that cites non-existent cases, or a tutoring system that presents wrong arithmetic can cause real harm. A model that scores well on standard Natural Language Understanding benchmarks can still hallucinate heavily in generation \cite{min2023factscore}. Without a comprehensive hallucination benchmark, there is no reliable way to assess how much an LLM can actually be trusted. As illustrated in Figure~\ref{fig:intro}, given a Bengali question about J.K. Rowling's mother and a supporting context stating the answer is \textit{Anne}, an LLM may instead generate \textit{Kathleen}-a fabricated fact that contradicts the provided context which BenHalluEval is designed to detect.

Hallucination evaluation has been explored in different languages such as English \cite{li2023halueval} Arabic \cite{alansari2025arahallueval}, Hindi and Turkish \cite{abdaljalil2026halluverse}. Bengali,  being the sixth most spoken language in the world \cite{sani2026addressingdatascarcitybangla}, remains underexplored. Dedicated pretrained models \cite{bhattacharjee2022banglabert}, multilingual benchmarks \cite{hasan2021xlsum}, and task-specific datasets \cite{clark2020tydiqa,khan2023banglachq,paul2025leveraging} already exist for Bengali, yet no benchmark measures hallucination in LLMs on both native Bengali and Bangla-English code-mixed input.

To fill this gap, we introduce \textbf{BenHalluEval}, a hallucination evaluation framework for Bengali covering four tasks: Question Answering (QA), Bangla-English Code-Mixed QA, Summarization, and Mathematical Reasoning. We use GPT-5.4\footnote{\url{https://openai.com/index/introducing-gpt-5-4/}} to generate hallucinated candidates, evaluate nine LLMs across three model categories, and systematically evaluate chain-of-thought prompting as a mitigation strategy across all model-task combinations.

\noindent Our main contributions are:
\begin{itemize}[leftmargin=*,noitemsep]
  \item We present \textbf{BenHalluEval}, a hallucination evaluation framework for Bengali spanning four generation tasks for LLMs on Bengali and Bangla-English inputs, covering Question Answering (QA), Bangla-English code-mixed QA, summarization, and mathematical reasoning.
  \item We construct 12,000 hallucinated candidates using GPT-5.4 across task-specific hallucination categories, including a Bangla-English code-mixed setting, and evaluate them alongside ground-truth instances using an LLM-as-judge framework.
  \item  We introduce \textbf{BenHalluScore} to jointly measure false positives and missed hallucinations, benchmark nine LLMs across three model categories, and systematically evaluate chain-of-thought prompting across all nine models and three tasks, showing the task-dependent patterns of improvement and degradation..
\end{itemize}

\section{Related Work}
\label{sec:related}

\noindent\textbf{Hallucination Evaluation Benchmarks.}
TruthfulQA, one of the earliest works on hallucination evaluation \cite{lin2022truthfulqa}, showed that model scale alone does not guarantee factual accuracy. FActScore advanced this by breaking long-form
outputs into atomic claims for fine-grained precision measurement \cite{min2023factscore}. HaluEval established the sampling-then-evaluation paradigm across QA, dialogue, and summarization  \cite{li2023halueval}. FaithBench \cite{bao2025faithbench}
and HalluLens \cite{hallulens2025} subsequently refined faithfulness annotations and guarded against benchmark leakage. MedHallu  extended this to the clinical domain, showing how domain specificity amplifies the impact of hallucination errors \cite{pandit2025medhallu}. A shared limitation across these benchmarks is labelling bias where LLM judges tend to default to one verdict class regardless of content, a problem none of them directly measure \cite{li2023halueval,adib2025assessing}.

\noindent\textbf{Low-Resource Hallucination and Bengali NLP.}
Hallucination rates scale inversely with pretraining data availability, with the gap widening in open-ended generation \cite{chataigner2024multilingual,islam2025multilingual}. AraHalluEval \cite{alansari2025arahallueval} evaluated Arabic-capable LLMs using fine-grained hallucination indicators, and Halluverse-M$^3$ \cite{abdaljalil2026halluverse} extended multilingual evaluation to English, Arabic, Hindi, and Turkish. Neither benchmark incorporates dual-track calibration or examines code-mixed input. More recent work covers multi-domain understanding via BnMMLU \cite{joy2025bnmmlu}, reasoning via
MathMist \cite{sobhani2025mathmist}, and summarization consistency via BanglaSummEval \cite{rafid2026banglasummeval}. Bangla-English code-mixed usage is well-documented \cite{alam2025bnsentmix,sheth2025codeswitched,alam2026mixsarc}, yet hallucination behaviour on Bangla-English input remains unexamined.
\paragraph{Positioning.} We state our novelty claim precisely against three nearby lines of work. First, Bengali LLM evaluation that does not address hallucination, such as BenLLMEval \cite{kabir2024benllm}. Second, hallucination benchmarks built for other individual languages, such as AraHalluEval \cite{alansari2025arahallueval}. Third, multilingual and fine-grained hallucination annotation benchmarks, such as ANAH-v2 \cite{gu2024anah}, HalluDial \cite{luo2024halludial}, and HalluVerse25 \cite{abdaljalil2025halluverse25}. Our claim is narrower than any of these: BenHalluEval is the first hallucination-\emph{detection} benchmark built specifically for Bengali that covers both native-script Bengali and Bangla-English code-mixed input, and evaluates them under a dual-track protocol.

\noindent\textbf{Mitigation Strategies.}
Chain-of-thought prompting \cite{wei2022cot} improves multi-step inference but can paradoxically obscure hallucination cues, making its benefit task- and model-dependent \cite{cheng2025chain}. \cite{cheng2025chain} show that CoT reasoning can shift a model's prediction patterns without improving its ability to distinguish hallucinated from correct content, with gains limited to tasks that require structured, verifiable intermediate steps.

BenHalluEval addresses this by covering four tasks: QA, Bangla-English Code-Mixed, QA, Summarization, and Mathematical Reasoning under a unified dual-track evaluation protocol.

\begin{figure*}[t]
  \centering
  \includegraphics[width=\textwidth]{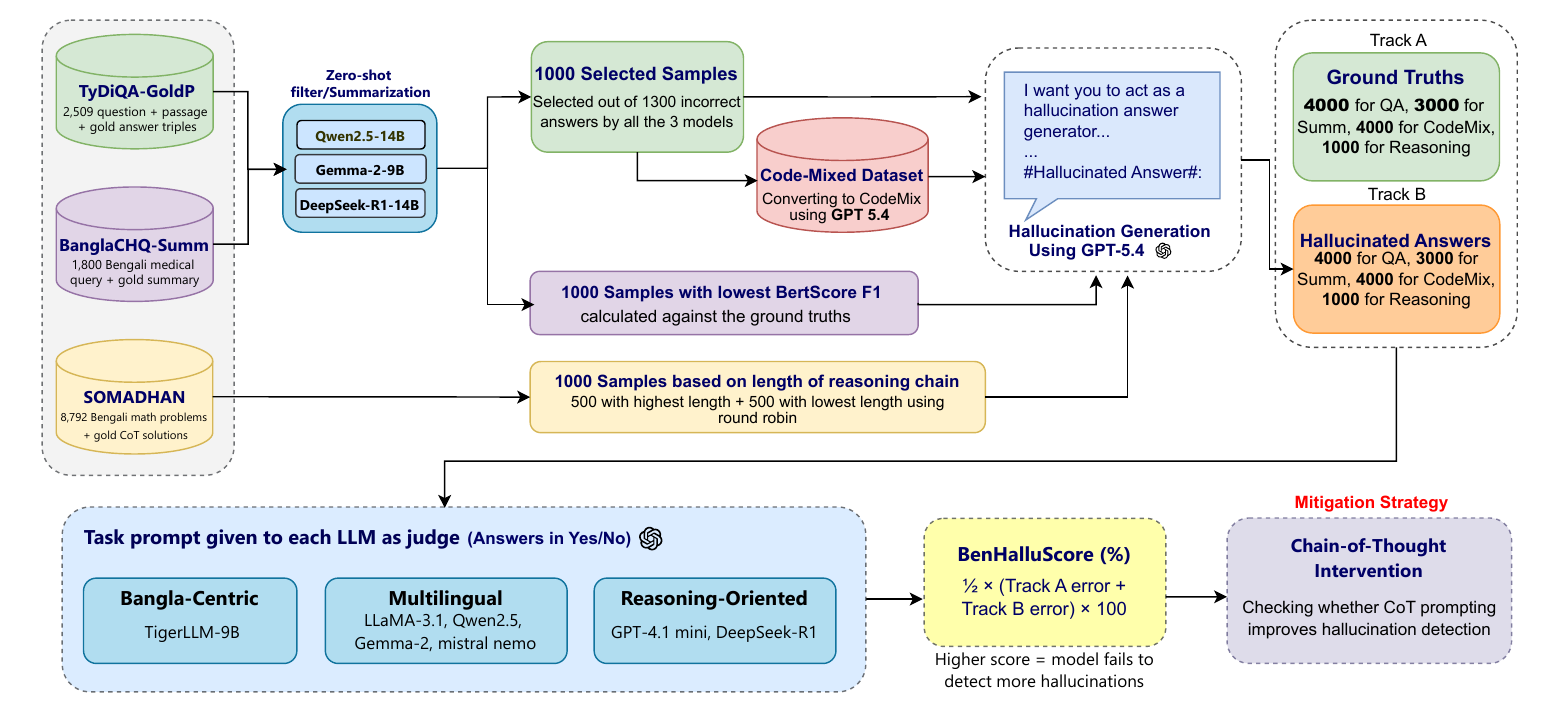}
  \caption{BenHalluEval system pipeline. Four tasks (QA, Code-Mixed QA, Summarization, Mathematical Reasoning) share a common evaluation stage. Track~A evaluates ground-truth correct answers; Track~B evaluates hallucinated candidates.} 
  \label{fig:pipeline}
\end{figure*}

\section{Methodology}
\label{sec:methodology}

BenHalluEval is constructed through a three-stage pipeline:
(i)~seed dataset selection and sample filtering,
(ii)~hallucinated sample generation using GPT-5.4, and
(iii)~hallucination detection evaluation using nine LLMs.
Figure~\ref{fig:pipeline} provides a full system overview.
Each task has a task-specific selection strategy and hallucination categorization; the evaluation protocol is unified across all four tasks.

\subsection{Stage 1: Seed Dataset Selection and Sample Filtering}

\paragraph{Context-based Question Answering.}
We use TyDiQA-GoldP (Bengali) \cite{clark2020tydiqa}, a dataset of 2,509 question--passage--answer triples. The supporting passage is provided to models at
evaluation time, so a hallucination verdict reflects consistency with the passage rather than open-domain world knowledge. To select the samples that are more prone to hallucination, we prompted three models zero-shot with only the question, without the passage: Qwen2.5-14B\footnote{\url{https://huggingface.co/Qwen/Qwen2.5-14B}}, Gemma-2-9B\footnote{\url{https://huggingface.co/google/gemma-2-9b}}, and DeepSeek-R1-14B\footnote{\url{https://ollama.com/library/deepseek-r1:14b}}. A Qwen2.5-32B-Instruct\footnote{\url{https://huggingface.co/Qwen/Qwen2.5-32B-Instruct}} judge scores each response as correct or incorrect. Samples that all three models answer incorrectly form the hard subset, yielding 1,313 instances (52.33\%), from which 1,000 are randomly selected as the QA seed dataset. These three models were chosen because they represent three distinct model families, fit within our available computational resources (a single RTX 3090 with 24,GB VRAM), and have been documented to perform competitively on Bengali tasks \cite{bhowmik2025evaluating,chae2026bengalimoralbench}. Since an instance is kept only if \emph{all three} fail, no single model's weakness determines the pool. To rule out same-family bias in the Qwen judge, we re-judged a random 500-instance subsample (seed 42) with GPT-4.1 mini: the judges agree on 95.2\%, 92.4\%, and 91.2\% of Qwen2.5-14B, DeepSeek-R1-14B, and Gemma-2-9B answers respectively, with disagreement \emph{lowest} for the same-family generator (Appendix~\ref{app:seed_analyses}). Table~\ref{tab:qa_filter} shows per-model zero-shot accuracy.

\begin{table}[h]
\centering
\small
\setlength{\tabcolsep}{10pt}
\begin{tabular}{lcc}
\toprule
\textbf{Model} & \textbf{Incorrect} & \textbf{Acc.\ (\%)} \\
\midrule
DeepSeek-R1-14B & 1,747 & 30.37 \\
Gemma-2-9B      & 1,668 & 33.54 \\
Qwen2.5-14B     & 1,948 & 22.36 \\
\midrule
All 3 incorrect & 1,313 & 52.33 \\
\bottomrule
\end{tabular}
\caption{Zero-shot performance  of three models on TyDiQA-GoldP (Bengali, 2,509 instances).}
\label{tab:qa_filter}
\end{table}
\paragraph{Bangla-English Code-Mixed QA.}
Bangla-English code-mixed text is widely used by Bengali speakers on social media and messaging platforms \cite{alam2025bnsentmix,sheth2025codeswitched,alam2026mixsarc}. To evaluate hallucination detection in this setting, we take the same 1,000 QA samples and convert the questions from Bengali to Bangla-English code-mixed using GPT-5.4 (see Appendix~\ref{app:conv_prompt}). Original answers are kept in
Bengali. No extra filtering is needed, since these samples are already the hard subset from the QA selection. To verify conversion quality, we manually reviewed 100 randomly selected samples. All authors are native Bengali speakers, and the reviewed samples were found to be natural and linguistically appropriate.

\paragraph{Summarization.}
We use BanglaCHQ-Summ \cite{khan2023banglachq}, a dataset of 1,800 Bengali consumer health query--summary pairs. We filter the samples based on semantic similarity using BERTScore F1. The same three models generate zero-shot summaries for all 1,800 samples, which are then scored against the original summaries using BERTScore F1 \cite{zhang2020bertscore}. The average BERTScore F1 is computed per sample across all three models. The 1,000 samples with the lowest average scores are selected, as these are the samples that all three models struggle most to summarize faithfully. We use this only as a Stage~1 difficulty filter, not a faithfulness signal. Appendix~\ref{app:nli} confirms that our seed selection criterion (BERTScore similarity) is independent of faithfulness (Spearman $\rho = -0.06$), while an NLI metric cleanly separates our hallucinated candidates from gold summaries (AUC~$= 0.90$), confirming the generated violations are genuine.

\paragraph{Mathematical Reasoning.}
We use SOMADHAN \cite{paul2025leveraging}, a dataset of 8,792 Bengali math word problems with manually written chain-of-thought (CoT) solutions. Samples are ranked by the length of their reasoning chain.The 500 longest and 500 shortest reasoning chains are selected as the two sampling pools. For each of the five error types, 100 samples with the longest reasoning chains and 100 with the shortest are selected, yielding 1,000 samples in total (200 per type). To ensure equal coverage, samples are picked alternately from the long and short pools in a fixed round-robin order, so that each error type receives an equal number of long and short reasoning chains. Chain length correlates with reasoning error patterns in large reasoning models \cite{zeng2025revisitingtesttimescalingo1like}; Appendix~\ref{app:chain_length} confirms this.

\subsection{Stage 2: Hallucinated Sample Generation}

All hallucinated samples are generated using GPT-5.4. Each prompt
specifies the hallucination type, includes an example of hallucination, and formats the target sample with field markers.

\paragraph{QA Hallucination Generation.}
Four hallucination types are applied to each of the 1,000 QA
seed samples, one API call per type (full prompt is in Appendix~\ref{app:gen_codemix_qa}). This generates 4,000
hallucinated candidates in total:

\begin{itemize}[leftmargin=*,noitemsep]
  \item \textbf{Factualness}: Fabricates a concrete fact such
    as a name, date, or place that contradicts the context.
  \item \textbf{Comprehension}: Misunderstands the question's
    context or intention and answers a different question.
  \item \textbf{Specificity}: Answers at an inappropriate level
    of detail, either too vague or too specific.
  \item \textbf{Inference}: Draws a conclusion that cannot be
    logically inferred from the given information.
\end{itemize}

\paragraph{Bangla-English Code-Mixed QA Hallucination Generation.}
The same four hallucination types used for QA are applied to
the 1,000 Bangla-English Code-mixed seed samples. The generation process mirrors the QA pipeline exactly, with one difference: both the prompt and the generated
hallucinated answers are in Bangla-English (Roman-script Bengali)
rather than native Bengali script (full prompt is in Appendix~\ref{app:gen_codemix_qa}).

\begin{table*}[t]
\centering
\small
\begin{tabularx}{\textwidth}{X}
\toprule
\rowcolor{promptblue}
I want you act as a hallucination answer generator. The answer should be given
in \textbf{BANGLA}. Given a question, right answer, and related knowledge, your
objective is to write a hallucinated answer that sounds plausible but is factually
incorrect. You SHOULD write the hallucinated answer using the following method: \\[3pt]
\rowcolor{promptpink}
1.\ You are trying to answer a question but there is a factual contradiction between
the answer and the knowledge. You can fabricate some information that does not exist
in the provided knowledge. \\[2pt]
\rowcolor{promptgreen}
\textit{$\langle$Demonstrations$\rangle$} \\[3pt]
\rowcolor{promptpink}
2.\ You are trying to answer a question but you misunderstand the question context
and intention. \\[2pt]
\rowcolor{promptgreen}
\textit{$\langle$Demonstrations$\rangle$} \\[3pt]
\rowcolor{promptpink}
3.\ You are trying to answer a question but the answer is too general or too specific
to answer the question at an appropriate level of specificity. \\[2pt]
\rowcolor{promptgreen}
\textit{$\langle$Demonstrations$\rangle$} \\[3pt]
\rowcolor{promptpink}
4.\ You are trying to answer a question but the answer cannot be inferred from the
knowledge. You can incorrectly reason with the knowledge to arrive at a hallucinated
answer. \\[2pt]
\rowcolor{promptgreen}
\textit{$\langle$Demonstrations$\rangle$} \\[4pt]
\rowcolor{promptyellow}
You should try your best to make the answer become hallucinated.
\texttt{\#Hallucinated Answer\#} can only have about 5 more words than
\texttt{\#Right Answer\#}. \\[4pt]
\rowcolor{promptyellow}
\textbf{\#Knowledge\#:} \textit{$\langle$insert the related knowledge/context$\rangle$} \\
\rowcolor{promptyellow}
\textbf{\#Question\#:} \textit{$\langle$insert the question$\rangle$} \\
\rowcolor{promptyellow}
\textbf{\#Right Answer\#:} \textit{$\langle$insert the right answer to the question$\rangle$} \\
\rowcolor{promptyellow}
\textbf{\#Hallucinated Answer\#:} Generate \\
\bottomrule
\end{tabularx}
\caption{Hallucination generation prompt for the QA task. Four methods correspond
to factual contradiction, comprehension mismatch, inappropriate specificity, and
incorrect inference errors. Each instance generates one hallucinated answer per
method under controlled constraints. Full prompt with Bengali demonstrations is
in Appendix~\ref{tab:prompt_qa_full}.}
\label{tab:prompt_qa}
\end{table*}

\paragraph{Summarization Hallucination Generation.}
Three hallucination types are applied to each of the 1,000
summarization seed instances, generating
3,000 hallucinated summaries in total:

\begin{itemize}[leftmargin=*,noitemsep]
  \item \textbf{Fabricated Content}: Introduces a detail that
    does not exist anywhere in the source document.
  \item \textbf{Non-Factual Addition}: Inserts a plausible but
    unverifiable claim not supported by the source.
  \item \textbf{Direct Contradiction}: Reverses information
    that is explicitly stated in the source.
\end{itemize}

Full prompt is in Appendix~\ref{app:gen_summ}.

\paragraph{Reasoning Hallucination Generation.}
We apply five error types, one per sample, selecting 
1,000 hallucinated CoT chains. GPT-5.4 is prompted to modify
exactly one reasoning step per instance and return output as
a structured JSON object (full prompt in
Appendix~\ref{app:gen_reason}):

\begin{itemize}[leftmargin=*,noitemsep]
  \item \textbf{Arithmetic Slip}: Silently alters a single
    numerical computation by a small amount.
  \item \textbf{Formula Misapplication}: Substitutes a related
    but incorrect mathematical operation.
  \item \textbf{Invalid Deduction}: Injects a logically
    unsupported transition between two reasoning steps.
  \item \textbf{Hallucinated Intermediate Fact}: Introduces a
    fabricated assumption mid-chain as if it were given.
  \item \textbf{Semantic Drift}: Reinterprets the question
    mid-reasoning so the final answer addresses a different goal.
\end{itemize}

Hallucination types are assigned by construction - one type per generation call which are adapted from HaluEval's taxonomy \cite{li2023halueval}. So labelling is single-label throughout. Category boundaries for conceptually adjacent types are detailed in
Appendix~\ref{app:hallu_types}.

\begin{table}[t]
\centering
\small
\setlength{\tabcolsep}{5pt}
\begin{tabular}{lcccc}
\toprule
\textbf{Task} & \textbf{Types} & \textbf{Seed} & \textbf{Hallu.} & \textbf{Gold} \\
\midrule
QA                    & 4 & 1,000 &  4,000 & 1,000 \\
Code-Mixed QA         & 4 & 1,000 &  4,000 & 1,000 \\
Summarization         & 3 & 1,000 &  3,000 & 1,000 \\
Math.\ Reasoning      & 5 & 1,000 &  1,000 & 1,000 \\
\midrule
Total                 &   &       & 12,000 & 4,000 \\
\bottomrule
\end{tabular}
\caption{BenHalluEval benchmark statistics.
``Hallu.''\ = hallucinated candidates generated by GPT-5.4;
``Gold''\ = ground-truth correct instances (Track~A evaluation).}
\label{tab:benchmark_stats}
\end{table}

\subsection{Stage 3: Hallucination Detection Evaluation}
\label{sec:eval_protocol}

\paragraph{Evaluation tracks.} Evaluating only on hallucinated candidates risks rewarding
models that blindly predict ``Yes'' for every input, while
evaluating only on ground-truth samples risks rewarding
models that always predict ``No''. To get a more accurate
picture of model behaviour, we evaluate each task on two
separate tracks. \textbf{Track~A} presents 1,000 ground-truth
correct answers, where the expected verdict is ``No'' (not
hallucinated). \textbf{Track~B} presents the hallucinated
candidates, where the expected verdict is ``Yes''. This
separation allows hallucination detection rate and
false-positive rate to be measured independently.

To measure performance across both tracks jointly, we introduce
\textbf{BenHalluScore} (BHS). A good model must both detect
hallucinated content and avoid flagging correct answers as
hallucinated. Neither track alone captures this. BenHalluScore
penalises both failure modes equally, so a higher score
indicates worse overall calibration. Formally:

\begin{equation}
    \text{BenHalluScore} = \frac{1}{2}
    \!\left(\frac{W_A}{N_A} + \frac{W_B}{N_B}\right)\! \times 100\%
\end{equation}

\paragraph{Relation to existing metrics.}
BenHalluScore equals Balanced Error Rate ($1 - \text{Balanced Accuracy}$): a model that answers ``Yes'' to everything scores 50\%, not the F$_1 = 0.89$ it would receive on our QA split. Probabilistic metrics (ECE, Brier score) do not apply since the protocol elicits a binary verdict, not a confidence estimate. Equal weighting is the unique symmetric setting under which neither degenerate responder wins. Appendix~\ref{app:weight_sweep} reports sensitivity to other weightings.

\paragraph{Human validation.} Three native Bengali-speaking annotators independently labelled
all the instances spanning all four tasks and twelve hallucination types
(Appendix~\ref{app:human_val}), blind to type and source, reaching almost-perfect agreement
(Fleiss' $\kappa = 0.911$--$0.926$; exact agreement 97--98\%). Code-mixed
conversions were validated separately for meaning preservation ($\kappa = 0.906$) and
naturalness (4.38/5).

where $W_A$ and $W_B$ are the number of wrong verdicts on
Track~A and Track~B respectively, $N_A$ and $N_B$ are the
total instances per track, and the factor $\frac{1}{2}$
normalises the combined error to the $[0, 100]$ range.
A score of 0\% indicates perfect calibration on both tracks;
50\% indicates random or uniformly biased behaviour;
100\% indicates perfectly wrong predictions across both tracks.

\begin{table*}[t]
\centering
\renewcommand{\arraystretch}{1.4}
\setlength{\tabcolsep}{5pt}
\resizebox{\textwidth}{!}{%
\begin{tabular}{l|ccc|ccc|ccc|ccc}
\toprule
\textbf{Model}
  & \multicolumn{3}{c|}{\textbf{QA}}
  & \multicolumn{3}{c|}{\textbf{Code-Mixed QA}}
  & \multicolumn{3}{c|}{\textbf{Summarization}}
  & \multicolumn{3}{c}{\textbf{Math.\ Reasoning}} \\
\cmidrule(lr){2-4}\cmidrule(lr){5-7}\cmidrule(lr){8-10}\cmidrule(lr){11-13}
& \textit{A-err.} & \textit{B-err.} & \textit{BHS}
& \textit{A-err.} & \textit{B-err.} & \textit{BHS}
& \textit{A-err.} & \textit{B-err.} & \textit{BHS}
& \textit{A-err.} & \textit{B-err.} & \textit{BHS} \\
\midrule
DeepSeek-R1-14B  & 48.00 & 29.20 & 38.60 & 40.10 & 13.78 & 26.94 & 17.60 &  6.73 & 12.17 & 19.90 & 16.30 & \cellcolor{cellbest}\textbf{18.10} \\
GPT-4.1 mini     & 13.90 & 17.23 & \cellcolor{cellbest}\textbf{15.56} & 24.00 & 19.80 & \cellcolor{cellbest}\textbf{21.90} &  8.20 & 11.37 &  9.79 & 31.70 & 21.10 & 26.40 \\
Qwen2.5-32B      & 27.90 & 65.80 & 46.85 & 17.40 & 30.60 & 24.00 &  4.70 & 10.73 & \cellcolor{cellbest}\textbf{7.72} & 24.80 & 28.80 & 26.80 \\
Gemma-2-27B      & 16.40 & 78.50 & 47.45 & 19.80 & 38.15 & 28.98 &  4.20 & 31.40 & 17.80 &  5.50 & 72.60 & 39.05 \\
Mistral-nemo-12B & 87.70 & 19.98 & \cellcolor{cellworst}\textbf{53.84} & 85.90 & 24.93 & \cellcolor{cellworst}\textbf{55.42} & 70.50 &  9.87 & 40.19 &  1.60 & 97.00 & 49.30 \\
LLaMA-3.1-8B     & 13.80 & 81.63 & 47.71 & 13.80 & 78.05 & 45.93 & 31.80 & 25.77 & 28.79 & 17.40 & 89.20 & 53.30 \\
TigerLLM-9B      & 18.40 & 38.17 & 28.28 & 25.80 & 37.28 & 31.54 &  4.80 & 47.53 & 26.17 & 21.40 & 85.50 & \cellcolor{cellworst}\textbf{53.45} \\
TituLLM-3B       & 99.75 &  0.25 & 50.00 & 99.25 &  2.00 & 50.62 & 100.00 &  0.33 & \cellcolor{cellworst}\textbf{50.16} & 100.00 &  0.00 & 50.00 \\
BanglaLLaMA-13B  & 83.25 & 14.00 & 48.62 & 85.50 & 11.25 & 48.38 & 89.00 &  4.67 & 46.84 & 100.00 &  0.00 & 50.00 \\
\bottomrule
\end{tabular}}
\caption{BenHalluScore results across all four tasks and nine models (all values in \%).
A-err.: Track~A error rate on ground-truth instances.
B-err.: Track~B error rate on hallucinated candidates.
\colorbox{cellbest}{\phantom{w}}~green cell = best (lowest) BHS per task;
\colorbox{cellworst}{\phantom{w}}~red cell = worst (highest) BHS per task.
Full per-task results with raw counts in Appendix~\ref{app:full_results}.}
\label{tab:combined_results}
\end{table*}

\section{Experimental Setup}
\label{sec:setup}

\paragraph{Models.}
We evaluate nine LLMs spanning three categories
(Table~\ref{tab:models}). The reasoning-oriented category comprises
DeepSeek-R1-14B and GPT-4.1 mini\footnote{\url{http://developers.openai.com/api/docs/models/gpt-4.1-mini}}.
The multilingual category includes Qwen2.5-32B-Instruct,
Gemma-2-27B\footnote{\url{https://huggingface.co/google/gemma-2-27b}},
Mistral-nemo-12B\footnote{\url{https://mistral.ai/news/mistral-nemo}}, and
LLaMA-3.1-8B-Instruct\footnote{\url{https://huggingface.co/meta-llama/Llama-3.1-8B-Instruct}}.
The Bangla-centric category comprises TigerLLM-9B \cite{raihan2025tigerllm},
TituLLM-3B \cite{nahin2025titullms}, and BanglaLLaMA-13B \cite{zehady2026banglallama}.
We deliberately include three systems here rather than one so that category-level statements
rest on more than a single model; all three are released with published Bengali benchmarking,
which lets us attribute differences within the category to factors other than exposure to
Bengali text.
Note that DeepSeek-R1-14B appears both in Stage~1 seed filtering and in the evaluated set.
It was \emph{not} used for hallucination generation, which was performed exclusively with
GPT-5.4, so no evaluated model is scored on candidates it authored; Stage~1 only determines
which seed instances enter the benchmark (Section~\ref{sec:methodology}).


\begin{table}[h]
\centering
\small
\begin{tabular}{llc}
\toprule
\textbf{Category} & \textbf{Model} & \textbf{Params} \\
\midrule
Reasoning       & DeepSeek-R1-14B       & 14B \\
                & GPT-4.1 mini          & ---  \\
\midrule
Multilingual    & Qwen2.5-32B-Instruct  & 32B \\
                & Gemma-2-27B           & 27B \\
                & Mistral-nemo-12B      & 12B \\
                & LLaMA-3.1-8B-Instruct & 8B  \\
\midrule
Bangla-centric  & TigerLLM-9B           & 9B  \\
                & TituLLM-3B            & 3B  \\
                & BanglaLLaMA-13B       & 13B \\
\bottomrule
\end{tabular}
\caption{Nine LLMs evaluated in BenHalluEval, grouped by
model category.}
\label{tab:models}
\end{table}

\paragraph{Inference Settings.}
All models produce binary verdicts (\textit{yes}/\textit{no})
at temperature~$= 0$, with a maximum output length of 1,024 tokens. Each instance is presented with its associated context - the passage for QA
and Bangla-English code-mixed QA, the source article for summarisation, and
the problem statement for reasoning. QA, summarisation, and
reasoning prompts are issued in Bengali and the code-mixed prompts
are in Bangla-English. GPT-4.1 mini is accessed via the
OpenAI API. All remaining models are served locally using
Ollama on a PC equipped with 128\,GB RAM and an
NVIDIA RTX\,3090 GPU (24\,GB VRAM). Full evaluation prompts
are provided in Appendix~\ref{app:eval_prompts}.

\section{Results Analysis}
\label{sec:results}

\subsection{Question Answering}

Table~\ref{tab:combined_results} shows QA results under the
dual-track BenHalluScore. GPT-4.1 mini achieves the best
BenHalluScore (15.56\%), maintaining low errors on both tracks
(A-err.\ 13.90\%, B-err.\ 17.23\%) - the only model that
performs well simultaneously on both. DeepSeek-R1-14B scores
38.60\%, with a notably high Track~A error of 48.00\%, meaning
it incorrectly flags nearly half of all correct answers as
hallucinated. Mistral-nemo-12B scores the worst (53.84\%),
driven almost entirely by its 87.70\% Track~A error - it flags
nearly every ground-truth answer as hallucinated. LLaMA-3.1-8B
shows the opposite pattern, predicting hallucination on 81.63\%
of Track~B instances while maintaining a low Track~A error
(13.80\%), reflecting opposing systematic biases rather than
content-sensitive judgements. Notably, TigerLLM-9B, the only Bengali-centric model evaluated, achieves the second-best BenHalluScore on QA
(28.28\%), outperforming significantly larger multilingual
models including Qwen2.5-32B, Gemma-2-27B, and even the
reasoning-oriented DeepSeek-R1-14B. This is not, however, a property of Bengali pretraining as such: the two other Bangla-centric models in our suite collapse to near-uniform verdicts on the same task.

\begin{table*}[!t]
\centering
\setlength{\tabcolsep}{6pt}
\renewcommand{\arraystretch}{1.1}
\begin{tabular}{ll|ccc||ccc}
\toprule
& & \multicolumn{3}{c||}{\textbf{Before CoT}}
  & \multicolumn{3}{c}{\textbf{After CoT}} \\
\cmidrule(lr){3-5}\cmidrule(lr){6-8}
\textbf{Model} & \textbf{Task}
  & \textbf{A-err.} & \textbf{B-err.} & \textbf{BHS}
  & \textbf{A-err.} & \textbf{B-err.} & \textbf{BHS} \\
\midrule
\multirow{4}{*}{DeepSeek-R1-14B}
  & QA            & 48.00 & 29.20 & 38.60 & 75.05 & 16.43 & \cellcolor{cellworst}\textbf{45.74}\,$\uparrow$ \\
  & Code-Mixed QA & 40.10 & 13.78 & 26.94 & 44.00 & 39.04 & \cellcolor{cellworst}\textbf{41.52}\,$\uparrow$ \\
  & Summarization & 17.60 &  6.73 & 12.17 & 20.70 &  9.07 & \cellcolor{cellworst}\textbf{14.89}\,$\uparrow$ \\
  & Math.\ Reasoning & 19.90 & 16.30 & 18.10 & 24.40 & 19.70 & \cellcolor{cellworst}\textbf{22.05}\,$\uparrow$ \\
\midrule
\multirow{4}{*}{GPT-4.1 mini}
  & QA            & 13.90 & 17.23 & 15.56 & 38.90 & 34.65 & \cellcolor{cellworst}\textbf{36.78}\,$\uparrow$ \\
  & Code-Mixed QA & 24.00 & 19.80 & 21.90 &  2.00 & 75.75 & \cellcolor{cellworst}\textbf{38.88}\,$\uparrow$ \\
  & Summarization &  8.20 & 11.37 &  9.79 &  6.30 & 11.80 & \cellcolor{cellbest}\textbf{9.05}\,$\downarrow$ \\
  & Math.\ Reasoning & 31.70 & 21.10 & 26.40 & 11.60 & 25.30 & \cellcolor{cellbest}\textbf{18.45}\,$\downarrow$ \\
\midrule
\multirow{4}{*}{Qwen2.5-32B}
  & QA            & 27.90 & 65.80 & 46.85 & 74.40 & 23.25 & \cellcolor{cellworst}\textbf{48.83}\,$\uparrow$ \\
  & Code-Mixed QA & 17.40 & 30.60 & 24.00 & 35.00 & 10.50 & \cellcolor{cellbest}\textbf{22.75}\,$\downarrow$ \\
  & Summarization &  4.70 & 10.73 &  7.72 & 33.60 &  7.47 & \cellcolor{cellworst}\textbf{20.54}\,$\uparrow$ \\
  & Math.\ Reasoning & 24.80 & 28.80 & 26.80 &  8.70 & 49.10 & \cellcolor{cellworst}\textbf{28.90}\,$\uparrow$ \\
\midrule
\multirow{4}{*}{Gemma-2-27B}
  & QA            & 16.40 & 78.50 & 47.45 & 50.50 & 38.58 & \cellcolor{cellbest}\textbf{44.54}\,$\downarrow$ \\
  & Code-Mixed QA & 19.80 & 38.15 & 28.98 & 51.00 & 46.25 & \cellcolor{cellworst}\textbf{48.62}\,$\uparrow$ \\
  & Summarization &  4.20 & 31.40 & 17.80 & 19.00 & 11.03 & \cellcolor{cellbest}\textbf{15.02}\,$\downarrow$ \\
  & Math.\ Reasoning &  5.50 & 72.60 & 39.05 &  3.60 & 82.30 & \cellcolor{cellworst}\textbf{42.95}\,$\uparrow$ \\
\midrule
\multirow{4}{*}{Mistral-nemo-12B}
  & QA            & 87.70 & 19.98 & 53.84 & 93.30 &  5.90 & \cellcolor{cellbest}\textbf{49.60}\,$\downarrow$ \\
  & Code-Mixed QA & 85.90 & 24.93 & 55.42 & 13.13 & 79.85 & \cellcolor{cellbest}\textbf{46.49}\,$\downarrow$ \\
  & Summarization & 70.50 &  9.87 & 40.19 & 45.40 & 14.10 & \cellcolor{cellbest}\textbf{29.75}\,$\downarrow$ \\
  & Math.\ Reasoning &  1.60 & 97.00 & 49.30 &  0.10 & 99.80 & \cellcolor{cellworst}\textbf{49.95}\,$\uparrow$ \\
\midrule
\multirow{4}{*}{LLaMA-3.1-8B}
  & QA            & 13.80 & 81.63 & 47.71 & 83.30 & 12.90 & \cellcolor{cellworst}\textbf{48.10}\,$\uparrow$ \\
  & Code-Mixed QA & 13.80 & 78.05 & 45.93 & 19.00 & 98.50 & \cellcolor{cellworst}\textbf{58.75}\,$\uparrow$ \\
  & Summarization & 31.80 & 25.77 & 28.79 & 64.90 & 21.80 & \cellcolor{cellworst}\textbf{43.35}\,$\uparrow$ \\
  & Math.\ Reasoning & 17.40 & 89.20 & 53.30 & 28.70 & 75.60 & \cellcolor{cellbest}\textbf{52.15}\,$\downarrow$ \\
\midrule
\multirow{4}{*}{TigerLLM-9B}
  & QA            & 18.40 & 38.17 & 28.28 & 21.80 & 38.37 & \cellcolor{cellworst}\textbf{30.09}\,$\uparrow$ \\
  & Code-Mixed QA & 25.80 & 37.28 & 31.54 &  0.00 & 99.50 & \cellcolor{cellworst}\textbf{49.75}\,$\uparrow$ \\
  & Summarization &  4.80 & 47.53 & 26.17 & 17.40 & 18.57 & \cellcolor{cellbest}\textbf{17.99}\,$\downarrow$ \\
  & Math.\ Reasoning & 21.40 & 85.50 & 53.45 &  8.40 & 52.00 & \cellcolor{cellbest}\textbf{30.20}\,$\downarrow$ \\
\midrule
\multirow{4}{*}{TituLLM-3B}
  & QA            &  99.75 & 0.25 & 50.00 & 63.50 & 39.25 & \cellcolor{cellworst}\textbf{51.38}\,$\uparrow$ \\
  & Code-Mixed QA &  99.25 & 2.00 & 50.62 & 34.00 & 65.99 & \cellcolor{cellbest}\textbf{50.00}\,$\downarrow$ \\
  & Summarization & 100.00 & 0.33 & 50.16 & 30.00 & 61.33 & \cellcolor{cellbest}\textbf{45.66}\,$\downarrow$ \\
  & Math.\ Reasoning & 100.00 & 0.00 & 50.00 & 66.00 & 31.00 & \cellcolor{cellbest}\textbf{48.50}\,$\downarrow$ \\
\midrule
\multirow{4}{*}{BanglaLLaMA-13B}
  & QA            &  83.25 & 14.00 & 48.62 & \multicolumn{3}{c}{N/A$^\ddagger$} \\
  & Code-Mixed QA &  85.50 & 11.25 & 48.38 & \multicolumn{3}{c}{N/A$^\ddagger$} \\
  & Summarization &  89.00 &  4.67 & 46.84 & \multicolumn{3}{c}{N/A$^\ddagger$} \\
  & Math.\ Reasoning & 100.00 & 0.00 & 50.00 & \multicolumn{3}{c}{N/A$^\ddagger$} \\
\bottomrule
\end{tabular}
\caption{%
  BenHalluScore~(BHS) before and after Chain-of-Thought~(CoT) prompting across all nine
  models and all four tasks.
  A-err.\ = false alarm rate (\%); B-err.\ = miss rate (\%);
  BHS~$= \tfrac{1}{2}(\text{A-err.}+\text{B-err.})$; lower is better.
  \colorbox{cellbest}{\phantom{x}}~improved ($\downarrow$);
  \colorbox{cellworst}{\phantom{x}}~degraded ($\uparrow$).
  $^\ddagger$BanglaLLaMA-13B produced no extractable yes/no verdict under CoT on any task,
  returning repetition-collapse output instead; we report N/A rather than a score, since a
  numeric value would invite the misreading that BHS $=50$ denotes random guessing when the
  underlying behaviour is systematic degeneration.
}
\label{tab:cot_results}
\end{table*}

\subsection{Bangla-English Code-Mixed QA}

Bangla-English code-mixed results differ from QA for several models
(Table~\ref{tab:combined_results}), showing that script
condition interacts with model behaviour. GPT-4.1 mini remains
the best (BHS 21.90\%), and Mistral-nemo-12B (55.42\%) and
LLaMA-3.1-8B (45.93\%) retain their characteristic biases
in the Roman-script condition. However, Qwen2.5-32B,
Gemma-2-27B, and DeepSeek-R1-14B improve sharply on Bangla-English code-mixed relative to QA (by 22.85, 18.47, and 11.66 BHS points
respectively), driven primarily by lower Track~A errors. A
plausible explanation is that Roman-script input shifts these
models toward their stronger English-anchored representations,
producing more consistent verdicts than native Bengali script -
the opposite of the intuitive expectation that code-mixing
would degrade performance.

\subsection{Summarization}

Across all models, BenHalluScores on summarization are notably lower 
than on QA (Table~\ref{tab:combined_results}), suggesting that access 
to a full source article provides stronger contextual basis for 
factual consistency judgements. Qwen2.5-32B achieves the lowest (best performing)
BenHalluScore (7.72\%), with low errors on both tracks (A-err.\ 4.70\%, B-err.\ 10.73\%), while DeepSeek-R1-14B ranks second
(12.17\%) despite recording the lowest Track~B error (6.73\%). Mistral-nemo-12B again scores worst (40.19\%), driven by a
70.50\% Track~A error. TigerLLM-9B shows the opposite pattern - a 47.53\% Track~B error against a 4.80\% Track~A error (BHS 26.17\%) - indicating that labelling bias persists 
even when the model has access to sufficient context.

\subsection{Mathematical Reasoning}

Mathematical reasoning is the hardest of the four tasks in the sense that matters here: its
median BenHalluScore is 49.30\%, so half the evaluated models perform at or worse than a
uniform responder that ignores content entirely (Table~\ref{tab:combined_results}). DeepSeek-R1-14B achieves the best and most balanced result (A-err.\ 19.90\%, B-err.\ 16.30\%, BHS 18.10\%). At the other end, Mistral-nemo-12B predicts ``No'' for nearly every instance regardless of content 
(A-err.\ 1.60\%, B-err.\ 97.00\%, BHS 49.30\%), a pattern indicating uniform label collapse. LLaMA-3.1-8B and TigerLLM-9B occupy the opposite extreme with Track~B errors of 89.20\% and 85.50\% respectively, resulting in the worst BenHalluScores (53.30\% and 53.45\%). The
concentration of models near or above 50\% BenHalluScore
indicates that reasoning hallucination in Bengali
is particularly difficult to detect.

\paragraph{Chain-of-Thought Mitigation.}
\label{sec:cot_results}
We apply CoT to all nine models across all four tasks. BanglaLLaMA-13B produced no
extractable verdict under CoT on any task, returning repetition-collapse output instead.
Its four zero-shot scores are retained in Table~\ref{tab:cot_results} but excluded from
the After-CoT analysis, leaving 32 evaluable After-CoT combinations across the remaining
eight models. CoT lowers BenHalluScore in 14 of these 32 and raises it in 18. Since
BenHalluScore is a balanced error rate, every reduction is by construction an improvement
in discrimination, but magnitude varies sharply. Only four combinations improve by more
than five points and finish at or below 30\% fac TigerLLM-9B Reasoning
(53.45$\rightarrow$30.20), Mistral-nemo-12B Summarization (40.19$\rightarrow$29.75),
TigerLLM-9B Summarization (26.17$\rightarrow$17.99), and GPT-4.1~mini Reasoning
(26.40$\rightarrow$18.45) --- and exactly one improves on \emph{both} tracks at once
(TigerLLM-9B Reasoning). Task governs the pattern more than model family: Summarization
improves for 5 of 8 models and Reasoning for 4 of 8, while QA degrades for 6 of 8 and
Code-Mixed QA for 5 of 8, consistent with \citet{cheng2025chain}'s finding that CoT
gains require verifiable intermediate steps. DeepSeek-R1-14B, which has built-in CoT,
degrades on all four tasks; GPT-4.1~mini, the strongest zero-shot QA model, suffers the
largest single degradation in the grid ($+21.22$ on QA).

\section{Discussion}
\label{sec:discussion}

\paragraph{Necessity of dual-track calibration.}
Detection-oriented benchmarks are conventionally scored on hallucinated candidates alone \cite{li2023halueval,gu2024anah,luo2024halludial,abdaljalil2025halluverse25}, leaving unexamined whether a model also recognises correct output as non-hallucinated. We note AraHalluEval~\cite{alansari2025arahallueval} evaluates output factuality rather than detection ability, so it is not directly comparable. Our contrast is with single-track
detection protocols. BenHalluEval's Track A exposes failures Track~B alone would conceal - Mistral-nemo-12B is competitive on Track B yet collapses on Track A - and the CoT results
reinforce this: several reductions in Track B error coincide with Track A error rising above 90\%, a trade BenHalluScore records as roughly neutral but a single-track metric would call progress.

\paragraph{QA vs.\ Code-Mixed QA.}
The Bangla-English Code-Mixed QA task shares its hallucination taxonomy and
candidate instances with QA, differing only in question script
(Bengali to Bangla-English). Despite this structural equivalence,
BenHalluScores diverge across both tasks
(Table~\ref{tab:combined_results}). Most multilingual models
achieve better results on Bangla-English Code-Mixed QA, whereas TigerLLM-9B and
GPT-4.1 mini perform more favourably on Bengali QA. This indicates
that the input script has a measurable influence on hallucination
judgement even when the underlying content is identical.

\paragraph{Bengali-centric pretraining is not sufficient on its own.}
Expanding the Bangla-centric category from one model to three does not support our earlier claim that Bengali pretraining explains TigerLLM-9B's calibration advantage: TituLLM-3B and BanglaLLaMA-13B, also Bangla-centric, sit at or near 50\% BenHalluScore on every task via
near-uniform verdicts (Table~\ref{tab:combined_results}). We therefore retract the stronger claim: parameter count is a poor predictor of Bengali hallucination calibration, but Bengali-centric pretraining is neither necessary nor sufficient for it - TigerLLM-9B's
advantage more plausibly reflects instruction tuning and prompt compatibility.

\paragraph{What separates strong from weak detectors.}
Three factors dominate. First, \textbf{verdict-format compliance}: TituLLM-3B and
BanglaLLaMA-13B score near 50\% by emitting a near-constant label rather than misjudging
content. Second, \textbf{evidence structure}: Summarization, with the longest explicit
evidence span, yields the lowest BenHalluScore for 8 of 9 models; Mathematical Reasoning,
requiring derivation rather than lookup, has the highest median (49.30\%). Third,
\textbf{script anchoring}: three multilingual models improve by up to 22.85 points on
Code-Mixed QA relative to QA despite identical content, while TigerLLM-9B and GPT-4.1~mini
move the other way. Parameter count predicts little.

\paragraph{Weighting robustness.}
We vary the relative weight of Track A and Track B errors across nine settings (Appendix~\ref{app:weight_sweep}). Model rankings stay stable in the middle range. Only at the extremes does a degenerate responder win - TituLLM-3B, which detects almost no hallucinations, tops three of four tasks when Track A is weighted at 0.1. Equal weighting is the only setting where no such responder can benefit.

\paragraph{Limits of prompting-based mitigation.}
CoT lowers BenHalluScore in 14 of 32 evaluable combinations and raises it in 18, governed more by task than model family: Summarization and Reasoning account for nine of the fourteen gains, while QA degrades for 6 of 8 models and Code-Mixed QA for 5 of 8, consistent with \citet{cheng2025chain}. Half of the gains land at or above 44\% BHS, so the model remains largely uninformative even after improving. Models with built-in CoT do not benefit from external CoT instruction: DeepSeek-R1-14B degrades on all four tasks, and the strongest zero-shot model, GPT-4.1~mini, suffers the largest single degradation in the grid.

\section{Conclusion}
\label{sec:conclusion}

Reliable hallucination evaluation for Bengali has remained absent despite the language's global reach. We introduced BenHalluEval, a benchmark spanning four Bengali and Bangla-English generation tasks with 12{,}000 hallucinated candidates and 4{,}000 gold instances, evaluated across nine LLMs under a dual-track protocol and validated by three
native-speaker annotators at almost-perfect agreement (Fleiss' $\kappa = 0.911$--$0.926$). We proposed BenHalluScore, a balanced dual-track measure that penalises false alarms on correct
output and missed hallucinations equally and so cannot be gamed by uniform response bias. BenHalluScore ranges from 7.72\% to 55.42\% across models and tasks, and rankings are task-conditional rather than scale-ordered: parameter count predicts calibration poorly, and
Bengali-centric pretraining is neither necessary nor sufficient for it. Chain-of-thought prompting lowers BenHalluScore in 14 of 32 evaluable combinations and raises it in 18, with
gains clustering on summarization and mathematical reasoning while both question-answering variants degrade predominantly. Prompting-only mitigation is therefore insufficient for Bengali.

\section*{Limitations}
While we conduct manual validation for selected components of the
benchmark, the hallucinated candidates are generated automatically
using GPT-5.4 rather than fully authored by human annotators.
As a result, the benchmark may still reflect generation-specific
patterns that differ from naturally occurring hallucinations in
deployed systems. We are extending candidate construction to a
second generator from a different family to quantify this.
BenHalluScore assigns equal weight to errors on
ground-truth and hallucinated tracks to reduce the influence of
systematic response tendencies, but deployment settings with
asymmetric costs may require task-specific weighting;
Appendix~\ref{app:weight_sweep} reports the full sensitivity
analysis. All models are evaluated zero-shot at temperature 0,
so results may differ under few-shot prompting or model-specific
prompt tuning. Two of the nine models (TituLLM-3B,
BanglaLLaMA-13B) were evaluated on a stratified sample rather
than the full benchmark, and BanglaLLaMA-13B could not be
evaluated under CoT at all. The chain-of-thought mitigation
experiment now covers all four tasks across eight models;
the interaction between code-mixed input and step-by-step
prompting is examined in Table~\ref{tab:cot_results}.
Finally, BenHalluEval is scoped to four Bengali and
Bangla-English task settings; generalization to other low-resource
languages, Bengali genres, domains, and more complex mathematical
reasoning remains to be validated.

\section*{Acknowledgements}
The authors acknowledge the use of ChatGPT for grammatical error 
checking only. All scientific content, analyses, and conclusions 
are solely the work of the authors.


\bibliography{references}

\appendix

\section{Annotation Protocol and Guidelines}
\label{app:annot_guidelines}
\paragraph{Annotators.} Three undergraduate annotators, all native Bengali speakers with
written and spoken Bangla coursework to at least the intermediate level. Annotators were
recruited from within the authors' institution, were informed of the research purpose, and
consented to the use of their annotations.

\paragraph{Task presented.} For each item the annotator saw the source context (passage,
document, or problem statement), the question or prompt where applicable, and one candidate
response, and answered a single binary question: \emph{does the response contain information
that is not supported by, or that contradicts, the given context?} Annotators did not see the
hallucination type, whether the item came from the gold or the generated pool, or any model
verdict. Item order was randomised and gold and hallucinated items were interleaved in
benchmark proportions.

\paragraph{Decision criteria.} (i) Judge only against the supplied context, never against
outside world knowledge. (ii) A response that is incomplete but contains nothing unsupported
is \emph{not} hallucinated. (iii) For summaries, an inference the document plainly licenses is
not hallucinated; an inference requiring an unstated premise is. (iv) For reasoning chains, a
chain is hallucinated if any step is arithmetically wrong, uses an unstated quantity, or
answers a different question than the one posed, even when the final answer is numerically
correct. (v) Fluency, register, and code-mixing style are never grounds for a hallucination
verdict.

\paragraph{Adjudication.} Ground truth was set by majority vote (2 of 3); no single annotator
could determine a label. Items on which an annotator abstained were excluded rather than
imputed; nine QA items were excluded on this basis.

\paragraph{Code-mixed conversion review.} The same annotators rated each sampled conversion on
two independent axes: meaning preservation relative to the Bengali original (binary) and
naturalness as Bangla--English written by a native speaker (1--5 Likert). Items unanimously
flagged as not preserving meaning were revised or excluded from the benchmark.

\section{Hallucination Type Definitions and Examples}
\label{app:hallu_types}

Table~\ref{tab:hallu_indicators} lists all hallucination types
used across the three generation tasks, with a definition and
a translated English example for each. Bangla-English Code-Mixed QA uses the same four types as QA, so we do not repeat them here.
\begin{table*}[h]
\centering
\small
\setlength{\tabcolsep}{4pt}
\begin{tabular}{
>{\raggedright\arraybackslash}m{2cm}
>{\raggedright\arraybackslash}p{2.2cm}
>{\raggedright\arraybackslash}p{4.5cm}
>{\raggedright\arraybackslash}p{4.5cm}
}
\toprule
\textbf{Task} & \textbf{Type} & \textbf{Definition} & \textbf{Example} \\
\midrule
\multirow{4}{*}{\textbf{QA}}
& Factualness
& Fabricates a concrete fact (name, date, number, place) not present in the context.
& \textbf{Q:} When was the Padma Bridge inaugurated? \newline
  \textbf{Ground-Truth:} 2022 \newline
  \textbf{Hallucinated:} The Padma Bridge was inaugurated in 2019. \\
\cmidrule{2-4}
& Comprehension
& Misunderstands the question context or intention.
& \textbf{Q:} What is the capital of Bangladesh? \newline
  \textbf{Ground-Truth:} Dhaka \newline
  \textbf{Hallucinated:} Bangladesh is a South Asian country. \\
\cmidrule{2-4}
& Specificity
& Answers at an inappropriate level of detail.
& \textbf{Q:} How many players are in a cricket team? \newline
  \textbf{Ground-Truth:} 11 \newline
  \textbf{Hallucinated:} Some. \\
\cmidrule{2-4}
& Inference
& Draws a conclusion not supported by the given information.
& \textbf{Q:} Where is Cox's Bazar located? \newline
  \textbf{Ground-Truth:} Chittagong Division \newline
  \textbf{Hallucinated:} Cox's Bazar is the capital of Bangladesh. \\
\midrule
\multirow{3}{*}{\textbf{Summ.}}
& Fabricated Content
& Introduces a detail not present anywhere in the source.
& \textbf{Context:} Patient reports mild fever for two days. \newline
  \textbf{Hallucinated:} Patient has been hospitalised for a week. \\
\cmidrule{2-4}
& Non-Factual Addition
& Inserts a plausible but unverifiable claim into the summary.
& \textbf{Context:} Child has a recurring cough. \newline
  \textbf{Hallucinated:} Child likely has bronchial asthma. \\
\cmidrule{2-4}
& Direct Contradiction
& Reverses information explicitly stated in the source.
& \textbf{Context:} Pain improves when walking. \newline
  \textbf{Hallucinated:} Pain does not improve with walking. \\
\midrule
\multirow{5}{*}{\textbf{Mathematical Reasoning}}
& Arithmetic Slip
& A single numerical computation is silently altered.
& \textbf{Correct:} $25 \times 8 = 200$ hours \newline
  \textbf{Hallucinated:} $25 \times 8 = 180$ hours \\
\cmidrule{2-4}
& Formula Misapplication
& A related but incorrect operation is substituted.
& \textbf{Correct:} $200 \times 5 = 1000$ \newline
  \textbf{Hallucinated:} $200 + 5 = 205$ \\
\cmidrule{2-4}
& Invalid Deduction
& A logically unsupported transition is injected between steps.
& \textbf{Correct:} Tax $= 20000 \times 0.1 = 2000$ \newline
  \textbf{Hallucinated:} Tax is split per pair of employees. \\
\cmidrule{2-4}
& Hallucinated Intermediate Fact
& A fabricated assumption is introduced mid-chain as if given.
& \textbf{Correct:} Working days $= 25$ \newline
  \textbf{Hallucinated:} Workers take 2 days off, so working days $= 23$. \\
\cmidrule{2-4}
& Semantic Drift
& The reasoning drifts to answer a different question than asked.
& \textbf{Correct:} Total cost $=$ wages $+$ tax \newline
  \textbf{Hallucinated:} Total $=$ wages only $=$ \$20,000. \\
\bottomrule
\end{tabular}
\caption{Hallucination type definitions and examples across QA,
Summarization, and Mathematical Reasoning. Bengali examples are translated to
English for readability.}
\label{tab:hallu_indicators}
\end{table*}

\paragraph{Labelling protocol.} Each candidate is produced under a prompt targeting exactly
one hallucination type (Appendix~\ref{app:gen_prompts}), so no multi-label adjudication
arises. Conceptually adjacent categories remain separable by their generative constraint:
\textit{Fabricated Content} introduces a detail with no referent in the source, whereas
\textit{Non-Factual Addition} elaborates an entity that \emph{is} present in the source with
an unverifiable claim; \textit{Arithmetic Slip} preserves the operator and perturbs only the
operand, whereas \textit{Formula Misapplication} preserves the operands and substitutes the
operator.

\paragraph{Why \textit{Inference} is a distinct QA category.} Because the supporting passage
is always provided, all four QA types are violations \emph{relative to that passage}; they
differ in how the violation arises. \textit{Factualness} requires the passage to state a
competing value for the same slot, which the hallucination replaces. \textit{Inference}
requires the passage to state \emph{no} value for that slot, and the hallucination supplies
one by an unlicensed step: a passage locating Cox's Bazar in Chittagong Division neither
states nor entails which city is the capital, so ``Cox's Bazar is the capital'' is
unsupported rather than contradicted. The operational test used during generation is whether
the hallucinated span could be deleted without changing the passage's truth conditions --- it
could not for \textit{Factualness}, it could for \textit{Inference}.

\paragraph{\textit{Semantic Drift} vs \textit{Hallucinated Intermediate Fact}.} A
\textit{Hallucinated Intermediate Fact} adds a new premise while leaving the goal untouched
(``workers take 2 days off, so working days $= 23$''). \textit{Semantic Drift} adds no
premise and changes the goal: every number used is one the problem supplied, but the chain
answers a quantity that was not asked for. Diagnostic: strike the hallucinated step and
re-run the chain. Under a hallucinated intermediate fact the chain recovers the correct
answer; under semantic drift it does not, because the target itself has moved.

\section{Benchmark Statistics by Hallucination Type}
\label{app:type_stats}
Table~\ref{tab:type_counts} reports the number of hallucinated
candidates generated per type and task, totalling 12,000
instances across the full benchmark.
\begin{table}[h]
\centering
\small
\begin{tabular}{llc}
\toprule
\textbf{Task} & \textbf{Hallucination type} & \textbf{Count} \\
\midrule
QA      & Factualness                    & 1,000 \\
         & Comprehension                  & 1,000 \\
         & Specificity                    & 1,000 \\
         & Inference                      & 1,000 \\
\midrule
Code-mix & Factualness                    & 1,000 \\
         & Comprehension                  & 1,000 \\
         & Specificity                    & 1,000 \\
         & Inference                      & 1,000 \\
\midrule
Summ.    & Fabricated content             & 1,000 \\
         & Non-factual addition           & 1,000 \\
         & Direct contradiction           & 1,000 \\
\midrule
Mathematical Reasoning & Arithmetic slip               & 200 \\
          & Formula misapplication        & 200 \\
          & Invalid deduction             & 200 \\
          & Hallucinated intermediate fact & 200 \\
          & Semantic drift                & 200 \\
\midrule
\multicolumn{2}{l}{\textbf{Total}} & \textbf{12,000} \\
\bottomrule
\end{tabular}
\caption{Hallucinated candidate counts by task and type.}
\label{tab:type_counts}
\end{table}

\section{Hallucination Generation Prompts}
\label{app:gen_prompts}

The following prompts were used to generate hallucinated
candidates via GPT-5.4. Each prompt specifies a hallucination
method, includes a demonstration, and formats the target
instance with field markers. Demonstrations are shown in
Bengali for QA and Mathematical Reasoning, and in Latin-script
Bangla-English for the Code-Mixed task.
 
\subsection{QA Generation Prompt}
Table~\ref{tab:prompt_qa_full} shows the full QA generation
prompt with Bengali demonstrations for all four methods.
 
\begin{table*}[h]
\centering\small
\begin{tabularx}{\textwidth}{X}
\toprule
\rowcolor{promptblue}
\textbf{[System instruction]} I want you act as a hallucination answer generator.
The answer should be given in \textbf{BANGLA}. Given a question, right answer, and
related knowledge, your objective is to write a hallucinated answer that sounds
plausible but is factually incorrect using one of the following methods: \\[2pt]
\rowcolor{promptpink}
\textbf{Method 1 --- Factualness:} You are trying to answer a question but there
is a factual contradiction between the answer and the knowledge. You can fabricate
some information that does not exist in the provided knowledge. \\[1pt]
\textbf{Demonstration:} \texttt{\#Knowledge\#:} ``{\bengalifont উইলিয়াম আব্রাহাম সাইমন ঔডারল্যান্ড (জন্ম: ৬ ডিসেম্বর, ১৯১৭ — মৃত্যু: ১৮ই মে, ২০০১) ছিলেন একজন ওলন্দাজ-অস্ট্রেলীয় সামরিক কমান্ডো অফিসার...}'' \texttt{\#Q\#:} ``{\bengalifont উইলিয়াম কবে জন্মগ্রহণ করেন?}'' \texttt{\#Gold\#:} ``{\bengalifont ৬ ডিসেম্বর, ১৯১৭}'' \texttt{\#Hallucinated\#:} ``{\bengalifont ৬ নভেম্বর, ১৯১৬}'' \\[2pt]
\rowcolor{promptpink}
\textbf{Method 2 --- Comprehension:} You are trying to answer a question but you
misunderstand the question context and intention. \\[1pt]
\textbf{Demonstration:} \texttt{\#Q\#:} ``{\bengalifont স্বাধীন বাংলাদেশের প্রথম চলচ্চিত্রটির নাম কী?}'' \texttt{\#Gold\#:} ``{\bengalifont সুকুমারী}'' \texttt{\#Hallucinated\#:} ``{\bengalifont জহির রায়হান}'' \\[2pt]
\rowcolor{promptpink}
\textbf{Method 3 --- Specificity:} You are trying to answer a question but the
answer is too general or too specific. \\[1pt]
\textbf{Demonstration:} \texttt{\#Q\#:} ``{\bengalifont কুয়েটের মোট ছাত্রছাত্রীর সংখ্যা কত?}'' \texttt{\#Gold\#:} ``{\bengalifont প্রায় ৬ হাজার}'' \texttt{\#Hallucinated\#:} ``{\bengalifont অজানা}'' \\[2pt]
\rowcolor{promptpink}
\textbf{Method 4 --- Inference:} You are trying to answer a question but the
answer cannot be inferred from the knowledge. \\[1pt]
\textbf{Demonstration:} \texttt{\#Q\#:} ``{\bengalifont ঢাকার মোট আয়তন কত?}'' \texttt{\#Gold\#:} ``{\bengalifont ১৩৪ বর্গমাইল}'' \texttt{\#Hallucinated\#:} ``{\bengalifont ২০ মিলিয়ন জনসংখ্যা}'' \\[2pt]
\rowcolor{promptyellow}
\texttt{\#Hallucinated Answer\#} can only have about 5 more words than \texttt{\#Right Answer\#}. \quad
\texttt{\#Knowledge\#:} $\langle$context$\rangle$ \quad \texttt{\#Question\#:} $\langle$question$\rangle$ \quad \texttt{\#Right Answer\#:} $\langle$gold$\rangle$ \quad \texttt{\#Hallucinated Answer\#:} Generate \\
\bottomrule
\end{tabularx}
\caption{QA hallucination generation prompt. One API call per method per seed instance.
The same structure is used for Code-Mixed QA with Bangla-English output and demonstrations.}
\label{tab:prompt_qa_full}
\end{table*}
 
\subsection{Code-Mixed QA Generation Prompt.}
\label{app:gen_codemix_qa}
Table~\ref{tab:prompt_gen_codemix_qa_full} mirrors the QA
prompt but instructs the model to generate hallucinated answers
in Latin-script Bangla-English throughout.
 
\begin{table*}[h]
\centering\small
\begin{tabularx}{\textwidth}{X}
\toprule
\rowcolor{promptblue}
\textbf{[System instruction]} I want you act as a hallucination answer generator.
The answer should be given in \textbf{CODE-MIXED BANGLA (Bangla-English: Latin-script Bengali)}.
Given a question, right answer, and related knowledge, write a hallucinated answer
that sounds plausible but is factually incorrect using the four methods below. \\[2pt]
\rowcolor{promptpink}
\textbf{Method 1 --- Factualness:} Fabricate information not present in the knowledge. \\[1pt]
\textbf{Demo:} \texttt{\#Knowledge\#:} ``William Abraham Simon Ouderland (janma: 6 december, 1917, mrittu: 18 may, 2001) chhilen ekjon commando officer.'' \texttt{\#Q\#:} ``Ouderland kobe janmogrohon koren?'' \texttt{\#Gold\#:} ``6 december, 1917'' \texttt{\#Hallucinated\#:} ``6 november, 1916'' \\[2pt]
\rowcolor{promptpink}
\textbf{Method 2 --- Comprehension:} Misunderstand the question context and intention. \\[1pt]
\textbf{Demo:} \texttt{\#Q\#:} ``Swadhin Bangladesher prothom cholocchitrotir nam ki?'' \texttt{\#Gold\#:} ``Sukumari'' \texttt{\#Hallucinated\#:} ``Jahir Raihan'' \\[2pt]
\rowcolor{promptpink}
\textbf{Method 3 --- Specificity:} Answer at an inappropriate level of specificity. \\[1pt]
\textbf{Demo:} \texttt{\#Q\#:} ``bortomane kuet-er mot chatrochhatrir shongkha koto?'' \texttt{\#Gold\#:} ``pray 6 hajar'' \texttt{\#Hallucinated\#:} ``ojana'' \\[2pt]
\rowcolor{promptpink}
\textbf{Method 4 --- Inference:} Draw a conclusion not supported by the knowledge. \\[1pt]
\textbf{Demo:} \texttt{\#Q\#:} ``Dhakar mot ayoton koto?'' \texttt{\#Gold\#:} ``134 borgomail'' \texttt{\#Hallucinated\#:} ``20 million jonoshonkha'' \\[2pt]
\rowcolor{promptyellow}
\texttt{\#Hallucinated Answer\#} can only have about 5 more words than \texttt{\#Right Answer\#}. \quad
\texttt{\#Knowledge\#:} $\langle$context$\rangle$ \quad \texttt{\#Question\#:} $\langle$question$\rangle$ \quad \texttt{\#Right Answer\#:} $\langle$gold$\rangle$ \quad \texttt{\#Hallucinated Answer\#:} Generate \\
\bottomrule
\end{tabularx}
\caption{Code-Mixed QA hallucination generation prompt. Four methods mirror the QA prompt
but use Latin-script Bangla-English throughout.}
\label{tab:prompt_gen_codemix_qa_full}
\end{table*}
 
\subsection{Summarization Generation Prompt.}
\label{app:gen_summ}
Table~\ref{tab:prompt_gen_summ} shows the summarization
generation prompt covering three hallucination methods.
 
\begin{table*}[h]
\centering\small
\begin{tabularx}{\textwidth}{X}
\toprule
\rowcolor{promptblue}
\textbf{[System instruction]} I want you act as a hallucination summary generator.
The summary should be given in \textbf{BANGLA}. Given a document and the right summary,
write a hallucinated summary that sounds plausible but is factually incorrect
using one of the following methods: \\[2pt]
\rowcolor{promptpink}
\textbf{Method 1 --- Fabricated Content:} You fabricate information that does not exist in the document. \\[1pt]
\textbf{Demo:} \texttt{\#Doc\#:} ``{\bengalifont আমার মেয়ের বয়স ৫ বছর। সর্দি সমস্যা।}'' \texttt{\#Gold\#:} ``{\bengalifont বয়স ৫, সর্দি সমস্যা, করণীয় কী}'' \texttt{\#Hallucinated\#:} ``{\bengalifont বয়স ৫, জ্বর ও কাশি সহ হাসপাতালে ভর্তি প্রয়োজন}'' \\[2pt]
\rowcolor{promptpink}
\textbf{Method 2 --- Non-Factual Addition:} Insert information that cannot be inferred from the document. \\[1pt]
\textbf{Demo:} \texttt{\#Doc\#:} ``{\bengalifont স্ত্রীর বয়স ২২, ডেট ৭ নভেম্বর থেকে ১৪ তারিখে পরিবর্তন।}'' \texttt{\#Gold\#:} ``{\bengalifont স্ত্রীর বয়স ২২, ডেট পরিবর্তন}'' \texttt{\#Hallucinated\#:} ``{\bengalifont ডাক্তার জুলাই মাসে ডেট দিয়ে পরে ১৪ করেছে}'' \\[2pt]
\rowcolor{promptpink}
\textbf{Method 3 --- Direct Contradiction:} Introduce a clear contradiction between summary and document. \\[1pt]
\textbf{Demo:} \texttt{\#Doc\#:} ``{\bengalifont হাঁটলে ব্যথা ঠিক হয়ে যায়।}'' \texttt{\#Gold\#:} ``{\bengalifont হাঁটলে ঠিক হয়}'' \texttt{\#Hallucinated\#:} ``{\bengalifont হাঁটলেও ব্যথা কমে না}'' \\[2pt]
\rowcolor{promptyellow}
\texttt{\#Hallucinated Summary\#} can only have about 5 more words than \texttt{\#Right Summary\#}. \quad
\texttt{\#Document\#:} $\langle$doc$\rangle$ \quad \texttt{\#Right Summary\#:} $\langle$gold$\rangle$ \quad \texttt{\#Hallucinated Summary\#:} Generate \\
\bottomrule
\end{tabularx}
\caption{Summarization hallucination generation prompt. Three methods correspond to
fabricated content, non-factual addition, and direct contradiction.}
\label{tab:prompt_gen_summ}
\end{table*}
 
\subsection{Reasoning Generation Prompt}
\label{app:gen_reason}
Table~\ref{tab:prompt_gen_reason} shows the reasoning
generation prompt. GPT-5.4 is instructed to modify exactly
one step and return output as a structured JSON object.
\begin{table*}[h]
\centering
\small
\begin{tabularx}{\textwidth}{X}
\toprule
\rowcolor{promptblue}
\textbf{[System instruction]} \\ I want you to act as a hallucinated reasoning chain
generator. The output must be in \textbf{BANGLA}. Given a math problem, a correct
step-by-step reasoning chain, and the correct final answer, your objective is to
produce a hallucinated reasoning chain that sounds plausible but is mathematically
incorrect. You MUST use one of the following hallucination methods: \\[3pt]
\rowcolor{promptpink}
\textbf{Method 1 — Arithmetic Slip:} \\ Keep the logic structure completely intact
but insert one subtle numerical error in a single calculation step. The wrong number
should be close to the correct one (roughly 5\% to 20\% off). All subsequent steps
must use the wrong number consistently. Preserve the \texttt{<<expr=result>>}
annotation format. \\[2pt]
\rowcolor{promptgreen}
\textbf{Demonstration:} \\
\texttt{\#Correct step\#:} ``{\bengalifont ২৫ দিন * ৮ ঘন্টা/দিন =
\textless\textless ২৫*৮=২০০\textgreater\textgreater ২০০ ঘন্টা}'' \\
\texttt{\#Hallucinated step\#:} ``{\bengalifont ২৫ দিন * ৮ ঘন্টা/দিন =
\textless\textless ২৫*৮=১৮০\textgreater\textgreater ১৮০ ঘন্টা}'' \\[3pt]
\rowcolor{promptpink}
\textbf{Method 2 — Formula Misapplication:} \\ Use a related but wrong operation
in one step. The Bengali narrative must justify the wrong operation naturally.
All subsequent steps must be consistent with the wrong result. \\[2pt]
\rowcolor{promptgreen}
\textbf{Demonstration:} \\
\texttt{\#Correct step\#:} ``{\bengalifont ২০০ ঘন্টা * ৳৫/ঘন্টা =
৳\textless\textless ২০০*৫=১০০০\textgreater\textgreater ১০০০}'' \\
\texttt{\#Hallucinated step\#:} ``{\bengalifont ২০০ + ৫ =
৳\textless\textless ২০০+৫=২০৫\textgreater\textgreater ২০৫}'' \\[3pt]
\rowcolor{promptpink}
\textbf{Method 3 — Invalid Deduction:} \\ Inject one logically unsupported
conclusion in a single step. All subsequent steps must follow from this
invalid conclusion. \\[2pt]
\rowcolor{promptgreen}
\textbf{Demonstration:} \\
\texttt{\#Correct step\#:} ``{\bengalifont ৳২০,০০০ * .১ =
৳\textless\textless ২০০০০*.১=২০০০\textgreater\textgreater ২,০০০}'' \\
\texttt{\#Hallucinated step\#:} ``{\bengalifont যেহেতু কর্মচারীর সংখ্যা জোড়
(৬ জন), তাই প্রতিটি জুটির জন্য আলাদাভাবে হিসাব করতে হবে।
প্রতি জুটির মজুরি = ৳২০,০০০ $\div$ ৩ =
\textless\textless ২০০০০/৩=৬৬৬৭\textgreater\textgreater ৳৬,৬৬৭।}'' \\[3pt]
\rowcolor{promptpink}
\textbf{Method 4 — Hallucinated Intermediate Fact:} \\ Introduce one fabricated
assumption not present in the problem as if it naturally follows. All subsequent
steps must proceed from this fabricated fact. \\[2pt]
\rowcolor{promptgreen}
\textbf{Demonstration:} \\
\texttt{\#Correct step\#:} ``{\bengalifont ২৫ দিন * ৮ ঘন্টা/দিন =
\textless\textless ২৫*৮=২০০\textgreater\textgreater ২০০ ঘন্টা}'' \\
\texttt{\#Hallucinated step\#:} ``{\bengalifont কর্মীরা প্রতি মাসে ২ দিন ছুটি
ভোগ করেন, তাই কার্যকর কাজের দিন = ২৫ - ২ =
\textless\textless ২৫-২=২৩\textgreater\textgreater ২৩ দিন।
মোট ঘন্টা = ২৩ * ৮ =
\textless\textless ২৩*৮=১৮৪\textgreater\textgreater ১৮৪ ঘন্টা।}'' \\[3pt]
\rowcolor{promptpink}
\textbf{Method 5 — Semantic Drift:} \\ Slightly reinterpret what the problem is
asking mid-reasoning so the final answer addresses a different question. The drift
must feel like a plausible reading of the problem. \\[2pt]
\rowcolor{promptgreen}
\textbf{Demonstration:} \\
\texttt{\#Correct step\#:} ``{\bengalifont ৳২,০০০ + ৳২০,০০০ =
৳\textless\textless ২০০০+২০০০০=২২০০০\textgreater\textgreater ২২,০০০}'' \\
\texttt{\#Hallucinated step\#:} ``{\bengalifont যেহেতু প্রশ্নটি মূলত মজুরির
মোট পরিমাণ জিজ্ঞেস করছে, তাই মোট মজুরি ৳২০,০০০ই চূড়ান্ত উত্তর।}'' \\[3pt]
\texttt{\#Problem\#:} \textit{$\langle$insert Bengali math problem$\rangle$} \\
\texttt{\#Correct Reasoning Chain\#:} \textit{$\langle$insert ground-truth CoT$\rangle$} \\
\texttt{\#Correct Answer\#:} \textit{$\langle$insert ground-truth answer$\rangle$} \\
\texttt{\#Hallucinated Chain\#:} Generate \\
\bottomrule
\end{tabularx}
\caption{Hallucination generation prompt for the Mathematical Reasoning task.
Five methods correspond to arithmetic slip, formula misapplication,
invalid deduction, hallucinated intermediate fact, and semantic drift.}
\label{tab:prompt_gen_reason}
\end{table*}


\subsection{Bengali-to-Bangla-English Conversion Prompt}
\label{app:conv_prompt}
Table~\ref{tab:prompt_conv} shows the Bengali-to-Bangla-English
conversion prompt used to produce the code-mixed seed instances
from the QA questions.
\begin{table*}[h]
\centering
\small
\begin{tabularx}{\textwidth}{X}
\toprule
\rowcolor{promptblue}
\textbf{[System instruction]} Convert Bengali QA text into natural Bangla-English —
Bengali--English code-mix written entirely in Latin script, the way people write
on South Asian social media or in casual chats. Output only Latin script - no
Bengali characters anywhere. Output only the converted text. No labels, no
explanations. \\[3pt]
\rowcolor{promptpink}
\textbf{Rules:} (1)~Romanise Bengali words naturally; use English words where
code-mix speakers naturally would. (2)~Do NOT translate everything to English;
preserve Bengali phrasing where it feels natural. (3)~Preserve meaning, tone,
names, numbers, dates, brands, and technical terms exactly. \\[3pt]
\rowcolor{promptgreen}
\textbf{Spelling variation} (apply to $\sim$15\% of eligible words): common
Bangla-English alternates: \textit{bhai/vai, bhalo/valo, ache/ase, ki/kee, onek/anek,
ektu/aktu, shob/sob, nai/nei, hoise/hoiche}. \\[2pt]
\rowcolor{promptyellow}
\textbf{Abbreviations} (apply sparingly, $\sim$10\% of eligible words):
\textit{phn} (phone), \textit{msg} (message), \textit{prob} (problem),
\textit{pls/plz} (please), \textit{bcz/coz} (because). \\[3pt]
\textbf{Do not:} over-abbreviate, use excessive expressive spellings, change
factual meaning, or make output hard to read. \\[4pt]
\texttt{[Text to convert]:} \textit{$\langle$Bengali question and/or passage$\rangle$} \\
\bottomrule
\end{tabularx}
\caption{Prompt used to convert the 1,000 QA seed instances from Bengali
to Bangla-English. Applied separately to questions and knowledge passages.
Ground-Truth answers are retained in their original Bengali form.}
\label{tab:prompt_conv}
\end{table*}

\section{Hallucination Evaluation Prompts}
\label{app:eval_prompts}
The following prompts were used to elicit binary hallucination
verdicts (\textit{yes}/\textit{no}) from the evaluated models.
Zero-shot prompts request a single-token response; chain-of-thought
prompts require step-by-step reasoning before the final verdict.
All prompts are in English except for the input fields, which
are in Bengali or Bangla-English depending on the task.

\subsection{Zero-Shot Evaluation Prompts}
Tables~\ref{tab:eval_qa}, \ref{tab:eval_summ}, and \ref{tab:eval_reason}
show the zero-shot evaluation prompts for QA/Code-Mixed QA,
Summarization, and Mathematical Reasoning respectively.

\begin{table}[h]
\centering
\small

\begin{tabularx}{\columnwidth}{X}
\toprule
\rowcolor{promptblue}
You are an evaluator. Decide whether the provided model answer is hallucinated
relative to the context and question. Only reply with a single token: \textbf{yes}
or \textbf{no}. No explanation, no punctuation, no extra text. \textit{yes} means
the answer contains information not supported by the context or contradicts it.
Provide the answer in English only. \\[4pt]
\texttt{Context:} \textit{$\langle$context$\rangle$} \\
\texttt{Question:} \textit{$\langle$question$\rangle$} \\
\texttt{Model answer:} \textit{$\langle$answer$\rangle$} \\[3pt]
\rowcolor{promptpink}
Answer now: \\
\bottomrule
\end{tabularx}
\caption{Zero-shot evaluation prompt for QA and Code-Mixed QA.
For code-mixed instances, all fields are in Bangla-English.}
\label{tab:eval_qa}

\vspace{1em}

\begin{tabularx}{\columnwidth}{X}
\toprule
\rowcolor{promptblue}
You are an evaluator. Decide whether the provided summary is hallucinated relative
to the document. Only reply with a single word: \textbf{yes} or \textbf{no}.
\textit{yes} means the summary contains information not supported by or contradicting
the document. Provide the answer in English only. \\[4pt]
\texttt{Document:} \textit{$\langle$Bengali medical article$\rangle$} \\
\texttt{Summary:} \textit{$\langle$candidate summary$\rangle$} \\[3pt]
\rowcolor{promptpink}
Answer now (yes or no): \\
\bottomrule
\end{tabularx}
\caption{Zero-shot evaluation prompt for Summarization.}
\label{tab:eval_summ}

\vspace{1em}

\begin{tabularx}{\columnwidth}{X}
\toprule
\rowcolor{promptblue}
You are an expert evaluator for Bengali mathematical reasoning tasks. Decide whether
the provided reasoning chain is hallucinated. Reply with a single token only:
\textbf{yes} or \textbf{no}. \textit{yes} = chain or answer contains incorrect
steps or unsupported facts. English only. \\[4pt]
\texttt{Problem:} \textit{$\langle$Bengali math problem$\rangle$} \\
\texttt{Reasoning Chain:} \textit{$\langle$CoT solution$\rangle$} \\
\texttt{Final Answer:} \textit{$\langle$final answer$\rangle$} \\[3pt]
\rowcolor{promptpink}
Answer now: \\
\bottomrule
\end{tabularx}
\caption{Zero-shot evaluation prompt for Mathematical Reasoning.}
\label{tab:eval_reason}

\end{table}

\subsection{Chain-of-Thought Evaluation Prompts}
Under the CoT condition, models are asked to reason through
three steps before issuing a verdict. Tables~\ref{tab:cot_qa},
\ref{tab:cot_summ}, and \ref{tab:cot_reason} show the
corresponding prompts for QA/Code-Mixed QA, Summarization,
and Mathematical Reasoning respectively.

\begin{table}[h]
\centering
\small

\begin{tabularx}{\columnwidth}{X}
\toprule
\rowcolor{promptblue}
You are an evaluator checking whether a model answer is hallucinated. \\[3pt]
\texttt{Question:} \textit{$\langle$question$\rangle$} \quad
\texttt{Model Answer:} \textit{$\langle$answer$\rangle$} \\[3pt]
\rowcolor{promptpink}
Analyze step by step: \\
\textbf{Step 1:} What factual claims does the answer make? \\
\textbf{Step 2:} Are these claims supported by or inferable from the question context? \\
\textbf{Step 3:} Based on steps 1--2, is the answer hallucinated? \\[3pt]
\rowcolor{promptgreen}
\textit{Final answer (write only this word on the last line):} \textbf{yes} or \textbf{no} \\
(\textit{yes} = hallucinated, \textit{no} = not hallucinated) \\
\bottomrule
\end{tabularx}
\caption{Chain-of-thought evaluation prompt for QA and Code-Mixed QA.}
\label{tab:cot_qa}

\vspace{1em}

\begin{tabularx}{\columnwidth}{X}
\toprule
\rowcolor{promptblue}
You are an evaluator checking whether a summary is hallucinated relative to a
document. \\[3pt]
\texttt{Document:} \textit{$\langle$document$\rangle$} \quad
\texttt{Summary:} \textit{$\langle$summary$\rangle$} \\[3pt]
\rowcolor{promptpink}
Analyze step by step: \\
\textbf{Step 1:} List the key claims made in the summary. \\
\textbf{Step 2:} For each claim, check whether it is directly supported by the document. \\
\textbf{Step 3:} Based on steps 1--2, decide your final answer. \\[3pt]
\rowcolor{promptgreen}
\textit{Final answer (write only this word on the last line):} \textbf{yes} or \textbf{no} \\
(\textit{yes} = hallucinated, \textit{no} = not hallucinated) \\
\bottomrule
\end{tabularx}
\caption{Chain-of-thought evaluation prompt for Summarization.}
\label{tab:cot_summ}

\vspace{1em}

\begin{tabularx}{\columnwidth}{X}
\toprule
\rowcolor{promptblue}
You are an expert evaluator for Bengali mathematical reasoning tasks. \\[3pt]
\texttt{Question:} \textit{$\langle$problem$\rangle$} \quad
\texttt{Reasoning Chain:} \textit{$\langle$CoT$\rangle$} \quad
\texttt{Answer:} \textit{$\langle$final answer$\rangle$} \\[3pt]
\rowcolor{promptpink}
Analyze step by step: \\
\textbf{Step 1:} Is each calculation or logical step mathematically correct? \\
\textbf{Step 2:} Does the final answer follow logically from the reasoning chain? \\
\textbf{Step 3:} Based on steps 1--2, is this reasoning chain hallucinated? \\[3pt]
\rowcolor{promptgreen}
Respond \textbf{ONLY} with a JSON object on the last line: \\
\texttt{\{"is\_hallucinated": "Yes"\}} or \texttt{\{"is\_hallucinated": "No"\}} \\
\bottomrule
\end{tabularx}
\caption{Chain-of-thought evaluation prompt for Mathematical Reasoning.}
\label{tab:cot_reason}

\end{table}
\section{Full Per-Task Evaluation Results}
\label{app:full_results}
The following tables extend Table~\ref{tab:combined_results}
in the main paper by reporting raw verdict counts alongside
error rates for all nine models across each of the four tasks.
A-err.: Track~A error rate on ground-truth instances.
B-err.: Track~B error rate on hallucinated candidates.
BHS: BenHalluScore $= \frac{1}{2}$(A-err.\ $+$ B-err.).

\subsection{Question Answering}

Table~\ref{tab:qa_results} shows the full QA evaluation results
across all nine models under the dual-track BenHalluScore.

\begin{table*}[h]
\centering
\small
\resizebox{\textwidth}{!}{%
\begin{tabular}{llccccccc}
\toprule
& & \multicolumn{3}{c}{\textbf{Track A --- Ground Truth (1,000)}}
  & \multicolumn{3}{c}{\textbf{Track B --- Hallucinated (4,000)}} & \\
\textbf{\#} & \textbf{Model} &
  \textbf{Correct} & \textbf{Wrong} & \textbf{A-err.} &
  \textbf{Correct} & \textbf{Wrong} & \textbf{B-err.} &
  \textbf{BHS} \\
\midrule
1 & Qwen2.5-32B-Instruct  & 771 & 279 & 27.90\% & 1,365 & 2,632 & 65.80\% & 46.85\% \\
2 & Gemma-2-27B            & 836 & 164 & 16.40\% &   860 & 3,140 & 78.50\% & 47.45\% \\
3 & DeepSeek-R1-14B        & 342 & 480 & 48.00\% & 2,820 & 1,168 & 29.20\% & 38.60\% \\
4 & GPT-4.1 mini           & 861 & 139 & 13.90\% & 3,311 &   689 & 17.23\% & \textbf{15.56\%} \\
5 & Mistral-nemo-12B       & 123 & 877 & 87.70\% & 3,201 &   799 & 19.98\% & \textbf{53.84\%} \\
6 & LLaMA-3.1-8B-Instruct  & 862 & 138 & 13.80\% &   735 & 3,265 & 81.63\% & 47.71\% \\
7 & TigerLLM-9B            & 816 & 184 & 18.40\% & 2,855 & 1,145 & 38.17\% & 28.28\% \\
\bottomrule
\end{tabular}}
\caption{QA full results. \textbf{Bold} = best and worst BHS.}
\label{tab:qa_results}
\end{table*}

\subsection{Summarization}

Table~\ref{tab:summ_results} shows the full summarization
evaluation results across all nine models under the dual-track
BenHalluScore.

\begin{table*}[h]
\centering
\small
\resizebox{\textwidth}{!}{%
\begin{tabular}{llccccccc}
\toprule
& & \multicolumn{3}{c}{\textbf{Track A --- Ground Truth (1,000)}}
  & \multicolumn{3}{c}{\textbf{Track B --- Hallucinated (3,000)}} & \\
\textbf{\#} & \textbf{Model} &
  \textbf{Correct} & \textbf{Wrong} & \textbf{A-err.} &
  \textbf{Correct} & \textbf{Wrong} & \textbf{B-err.} &
  \textbf{BHS} \\
\midrule
1 & Qwen2.5-32B-Instruct  & 953 &  47 &  4.70\% & 2,678 &   322 & 10.73\% & \textbf{7.72\%} \\
2 & Gemma-2-27B            & 958 &  42 &  4.20\% & 2,058 &   942 & 31.40\% & 17.80\% \\
3 & DeepSeek-R1-14B        & 824 & 176 & 17.60\% & 2,798 &   202 &  6.73\% & 12.17\% \\
4 & GPT-4.1 mini           & 918 &  82 &  8.20\% & 2,659 &   341 & 11.37\% &  9.79\% \\
5 & Mistral-nemo-12B       & 295 & 705 & 70.50\% & 2,704 &   296 &  9.87\% & \textbf{40.19\%} \\
6 & LLaMA-3.1-8B-Instruct  & 682 & 318 & 31.80\% & 2,227 &   773 & 25.77\% & 28.79\% \\
7 & TigerLLM-9B            & 952 &  48 &  4.80\% & 1,574 & 1,426 & 47.53\% & 26.17\% \\
\bottomrule
\end{tabular}}
\caption{Summarization full results. \textbf{Bold} = best and worst BHS.}
\label{tab:summ_results}
\end{table*}

\subsection{Mathematical Reasoning}

Table~\ref{tab:reason_results} shows the full mathematical
reasoning evaluation results across all nine models under
the dual-track BenHalluScore.

\begin{table*}[h]
\centering
\small
\resizebox{\textwidth}{!}{%
\begin{tabular}{llccccccc}
\toprule
& & \multicolumn{3}{c}{\textbf{Track A --- Ground Truth (1,000)}}
  & \multicolumn{3}{c}{\textbf{Track B --- Hallucinated (1,000)}} & \\
\textbf{\#} & \textbf{Model} &
  \textbf{Correct} & \textbf{Wrong} & \textbf{A-err.} &
  \textbf{Correct} & \textbf{Wrong} & \textbf{B-err.} &
  \textbf{BHS} \\
\midrule
1 & Qwen2.5-32B-Instruct  & 752 & 248 & 24.80\% & 703 & 288 & 28.80\% & 26.80\% \\
2 & Gemma-2-27B            & 945 &  55 &  5.50\% & 274 & 726 & 72.60\% & 39.05\% \\
3 & DeepSeek-R1-14B        & 801 & 199 & 19.90\% & 837 & 163 & 16.30\% & \textbf{18.10\%} \\
4 & GPT-4.1 mini           & 683 & 317 & 31.70\% & 789 & 211 & 21.10\% & 26.40\% \\
5 & Mistral-nemo-12B       & 984 &  16 &  1.60\% &  30 & 970 & 97.00\% & 49.30\% \\
6 & LLaMA-3.1-8B-Instruct  & 826 & 174 & 17.40\% & 108 & 892 & 89.20\% & 53.30\% \\
7 & TigerLLM-9B            & 786 & 214 & 21.40\% & 145 & 855 & 85.50\% & \textbf{53.45\%} \\
\bottomrule
\end{tabular}}
\caption{Mathematical Reasoning full results. \textbf{Bold} = best and worst BHS.}
\label{tab:reason_results}
\end{table*}

\subsection{Bangla-English Code-Mixed QA}

Table~\ref{tab:codemix_results} shows the full Code-Mixed
Banglish QA evaluation results across all nine models under
the dual-track BenHalluScore.

\begin{table*}[h]
\centering
\small
\resizebox{\textwidth}{!}{%
\begin{tabular}{llccccccc}
\toprule
& & \multicolumn{3}{c}{\textbf{Track A --- Ground Truth (1,000)}}
  & \multicolumn{3}{c}{\textbf{Track B --- Hallucinated (4,000)}} & \\
\textbf{\#} & \textbf{Model} &
  \textbf{Correct} & \textbf{Wrong} & \textbf{A-err.} &
  \textbf{Correct} & \textbf{Wrong} & \textbf{B-err.} &
  \textbf{BHS} \\
\midrule
1 & Qwen2.5-32B-Instruct  & 826 & 174 & 17.40\% & 2,776 & 1,224 & 30.60\% & 24.00\% \\
2 & Gemma-2-27B            & 802 & 198 & 19.80\% & 2,474 & 1,526 & 38.15\% & 28.98\% \\
3 & DeepSeek-R1-14B        & 596 & 401 & 40.10\% & 3,439 &   551 & 13.78\% & 26.94\% \\
4 & GPT-4.1 mini           & 760 & 240 & 24.00\% & 3,208 &   792 & 19.80\% & \textbf{21.90\%} \\
5 & Mistral-nemo-12B       & 141 & 859 & 85.90\% & 3,003 &   997 & 24.93\% & \textbf{55.42\%} \\
6 & LLaMA-3.1-8B-Instruct  & 862 & 138 & 13.80\% &   878 & 3,122 & 78.05\% & 45.93\% \\
7 & TigerLLM-9B            & 739 & 258 & 25.80\% & 2,496 & 1,491 & 37.28\% & 31.54\% \\
\bottomrule
\end{tabular}}
\caption{Bangla-English Code-Mixed QA full results. \textbf{Bold} = best and worst BHS.}
\label{tab:codemix_results}
\end{table*}

\subsection{QA Model Error Distribution}
Table~\ref{tab:qa_results} reports full per-model Track~A and
Track~B error rates on QA. GPT-4.1 mini achieves the best balance
(BHS 15.56\%), while Mistral-nemo-12B scores worst (BHS 53.84\%),
driven by its 87.70\% Track~A error. Models such as LLaMA-3.1-8B
show the opposite pattern, with high Track~B error and low Track~A
error, reflecting a systematic ``Yes'' bias rather than genuine
hallucination discrimination.

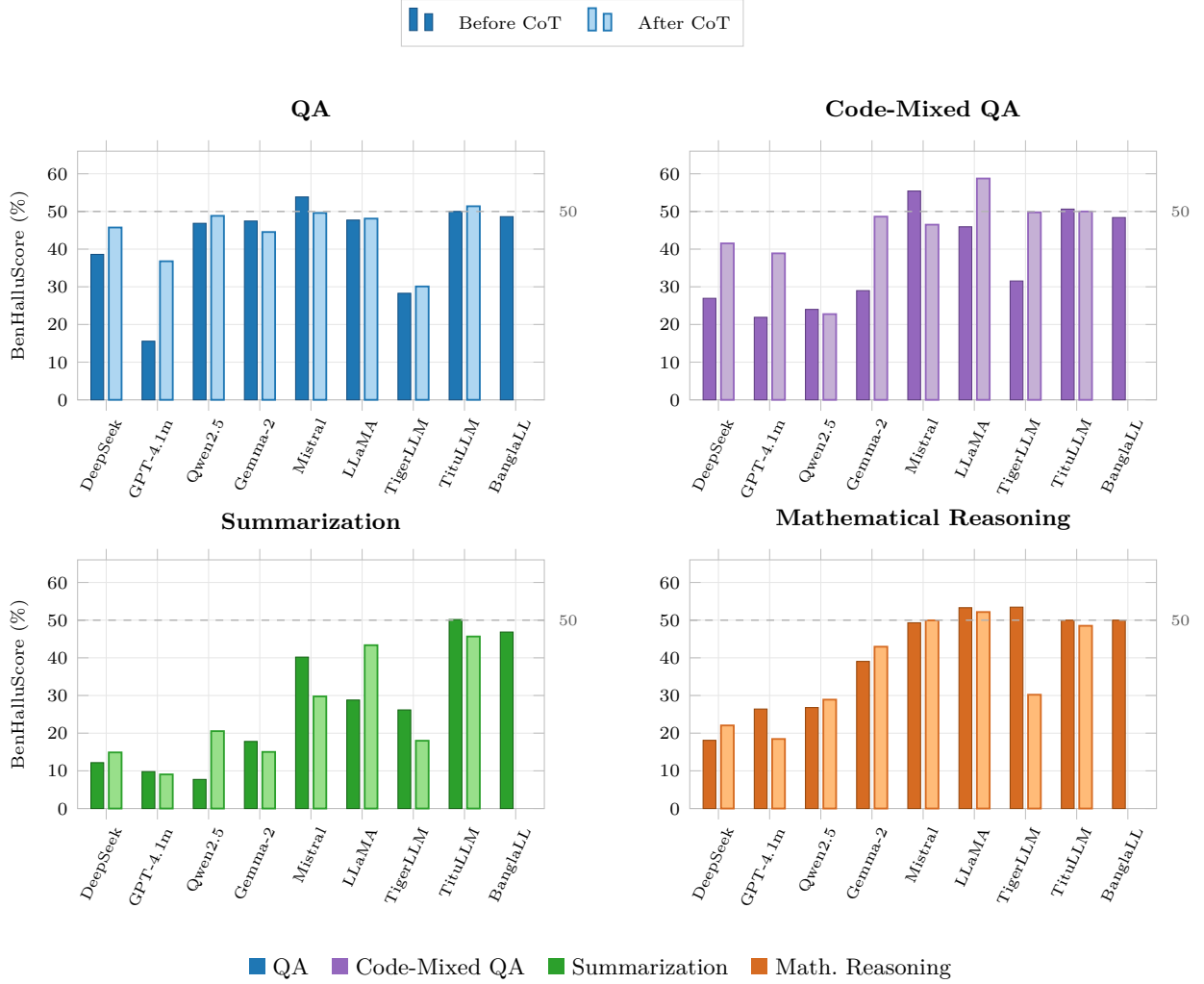
\begin{figure*}[t]
\centering
\begin{tikzpicture}
\begin{groupplot}[
    group style={
        group size=2 by 2,
        horizontal sep=2.0cm,
        vertical sep=2.2cm,
    },
    ybar,
    width=0.50\textwidth,
    height=5.0cm,
    ymin=0, ymax=66,
    ytick={0,10,20,30,40,50,60},
    yticklabel style={font=\scriptsize},
    ylabel style={font=\scriptsize},
    symbolic x coords={
        DeepSeek, GPT-4.1m, Qwen2.5, Gemma-2,
        Mistral, LLaMA, TigerLLM, TituLLM, BanglaLL},
    xtick=data,
    xticklabel style={font=\tiny, rotate=65, anchor=north east,
                      inner sep=1pt, yshift=-2pt},
    xtick align=outside,
    grid=major,
    grid style={line width=0.3pt, draw=gray!20},
    axis line style={gray!50},
    tick style={gray!50},
    enlarge x limits=0.07,
    title style={font=\small\bfseries, yshift=2pt},
    clip=false,
]

\nextgroupplot[
    bar width=5pt,
    title={QA},
    ylabel={BenHalluScore (\%)},
    legend style={
        font=\scriptsize,
        at={(1.06, 1.42)},
        anchor=south,
        legend columns=2,
        column sep=8pt,
        inner sep=3pt,
        draw=gray!40,
    }
]
\addplot[fill={rgb,255:red,31;green,119;blue,180},
         draw={rgb,255:red,20;green,80;blue,130}, line width=0.4pt]
  coordinates{
    (DeepSeek,38.60)(GPT-4.1m,15.56)(Qwen2.5,46.85)(Gemma-2,47.45)
    (Mistral,53.84)(LLaMA,47.71)(TigerLLM,28.28)(TituLLM,50.00)(BanglaLL,48.62)};
\addlegendentry{Before CoT}
\addplot[fill={rgb,255:red,174;green,214;blue,241},
         draw={rgb,255:red,31;green,119;blue,180}, line width=0.7pt]
  coordinates{
    (DeepSeek,45.74)(GPT-4.1m,36.78)(Qwen2.5,48.83)(Gemma-2,44.54)
    (Mistral,49.60)(LLaMA,48.10)(TigerLLM,30.09)(TituLLM,51.38)};
\addlegendentry{After CoT}
\draw[dashed, gray!60, line width=0.55pt]
  ({rel axis cs:0,0}|-{axis cs:DeepSeek,50})--
  ({rel axis cs:1,0}|-{axis cs:DeepSeek,50});
\node[font=\tiny, gray!70!black, anchor=west]
  at ({rel axis cs:1.01,0}|-{axis cs:DeepSeek,50}) {50};

\nextgroupplot[bar width=5pt, title={Code-Mixed QA}]
\addplot[fill={rgb,255:red,148;green,103;blue,189},
         draw={rgb,255:red,95;green,60;blue,130}, line width=0.4pt]
  coordinates{
    (DeepSeek,26.94)(GPT-4.1m,21.90)(Qwen2.5,24.00)(Gemma-2,28.98)
    (Mistral,55.42)(LLaMA,45.93)(TigerLLM,31.54)(TituLLM,50.62)(BanglaLL,48.38)};
\addplot[fill={rgb,255:red,197;green,176;blue,213},
         draw={rgb,255:red,148;green,103;blue,189}, line width=0.7pt]
  coordinates{
    (DeepSeek,41.52)(GPT-4.1m,38.88)(Qwen2.5,22.75)(Gemma-2,48.62)
    (Mistral,46.49)(LLaMA,58.75)(TigerLLM,49.75)(TituLLM,50.00)};
\draw[dashed, gray!60, line width=0.55pt]
  ({rel axis cs:0,0}|-{axis cs:DeepSeek,50})--
  ({rel axis cs:1,0}|-{axis cs:DeepSeek,50});
\node[font=\tiny, gray!70!black, anchor=west]
  at ({rel axis cs:1.01,0}|-{axis cs:DeepSeek,50}) {50};

\nextgroupplot[bar width=5pt, title={Summarization}, ylabel={BenHalluScore (\%)}]
\addplot[fill={rgb,255:red,44;green,160;blue,44},
         draw={rgb,255:red,20;green,100;blue,20}, line width=0.4pt]
  coordinates{
    (DeepSeek,12.17)(GPT-4.1m,9.79)(Qwen2.5,7.72)(Gemma-2,17.80)
    (Mistral,40.19)(LLaMA,28.79)(TigerLLM,26.17)(TituLLM,50.16)(BanglaLL,46.84)};
\addplot[fill={rgb,255:red,152;green,223;blue,138},
         draw={rgb,255:red,44;green,160;blue,44}, line width=0.7pt]
  coordinates{
    (DeepSeek,14.89)(GPT-4.1m,9.05)(Qwen2.5,20.54)(Gemma-2,15.02)
    (Mistral,29.75)(LLaMA,43.35)(TigerLLM,17.99)(TituLLM,45.66)};
\draw[dashed, gray!60, line width=0.55pt]
  ({rel axis cs:0,0}|-{axis cs:DeepSeek,50})--
  ({rel axis cs:1,0}|-{axis cs:DeepSeek,50});
\node[font=\tiny, gray!70!black, anchor=west]
  at ({rel axis cs:1.01,0}|-{axis cs:DeepSeek,50}) {50};

\nextgroupplot[bar width=5pt, title={Mathematical Reasoning}]
\addplot[fill={rgb,255:red,214;green,107;blue,34},
         draw={rgb,255:red,150;green,70;blue,10}, line width=0.4pt]
  coordinates{
    (DeepSeek,18.10)(GPT-4.1m,26.40)(Qwen2.5,26.80)(Gemma-2,39.05)
    (Mistral,49.30)(LLaMA,53.30)(TigerLLM,53.45)(TituLLM,50.00)(BanglaLL,50.00)};
\addplot[fill={rgb,255:red,255;green,187;blue,120},
         draw={rgb,255:red,214;green,107;blue,34}, line width=0.7pt]
  coordinates{
    (DeepSeek,22.05)(GPT-4.1m,18.45)(Qwen2.5,28.90)(Gemma-2,42.95)
    (Mistral,49.95)(LLaMA,52.15)(TigerLLM,30.20)(TituLLM,48.50)};
\draw[dashed, gray!60, line width=0.55pt]
  ({rel axis cs:0,0}|-{axis cs:DeepSeek,50})--
  ({rel axis cs:1,0}|-{axis cs:DeepSeek,50});
\node[font=\tiny, gray!70!black, anchor=west]
  at ({rel axis cs:1.01,0}|-{axis cs:DeepSeek,50}) {50};

\end{groupplot}

\node[anchor=north, yshift=-0.5cm, font=\small]
  at (current bounding box.south) {%
  \tikz\fill[fill={rgb,255:red,31;green,119;blue,180}]
    (0,0) rectangle (0.22,0.22);~QA\quad
  \tikz\fill[fill={rgb,255:red,148;green,103;blue,189}]
    (0,0) rectangle (0.22,0.22);~Code-Mixed QA\quad
  \tikz\fill[fill={rgb,255:red,44;green,160;blue,44}]
    (0,0) rectangle (0.22,0.22);~Summarization\quad
  \tikz\fill[fill={rgb,255:red,214;green,107;blue,34}]
    (0,0) rectangle (0.22,0.22);~Math.\ Reasoning
};

\end{tikzpicture}
\caption{BenHalluScore before and after chain-of-thought prompting across all nine models
and all four tasks (32 evaluable combinations; BanglaLLaMA-13B has no \emph{After} bar on
any task, having produced only repetition-collapse output under CoT). Each panel shows one
task; darker bars = Before CoT, lighter bars = After CoT. Lower is better. The dashed line
at 50\% marks the score attained by a uniform responder that ignores content; reductions
that still land near this line represent a redistribution of response bias rather than a
move to a useful operating point (Section~\ref{sec:cot_results}).}
\label{fig:cot_bhs}
\end{figure*}

\section{BenHalluScore Weight Sensitivity}
\label{app:weight_sweep}
We recompute $\text{BHS}_{w} = w_A\,(\text{A-err.}) + (1-w_A)\,(\text{B-err.})$ sweeping $w_A \in \{0.1, \dots, 0.9\}$ for all nine models and all four tasks. Among models that produce content-sensitive verdicts, rankings are highly stable: GPT-4.1 mini tops QA at every weighting from $w_A = 0.2$ to $0.9$, and DeepSeek-R1-14B holds the top rank on Mathematical Reasoning from $w_A = 0.2$ through $0.7$. At $w_A = 0.1$ the nominal winner is TituLLM-3B on QA, Code-Mixed QA and Mathematical Reasoning - a model that answers ``No'' on 99-100\% of gold instances - and at $w_A = 0.9$ on Mathematical Reasoning, Mistral-nemo-12B (A-err.\ 1.60\%, B-err.\ 97.00\%) scores 11.1, below DeepSeek-R1-14B's 19.5, despite flagging almost none of the 1{,}000 hallucinated chains it was shown.

\begin{table*}[h]
\centering\small
\setlength{\tabcolsep}{4pt}
\begin{tabular}{ll ccccccccc}
\toprule
\textbf{Task} & \textbf{Model} & \multicolumn{9}{c}{$w_A$} \\
\cmidrule(lr){3-11}
 & & 0.1 & 0.2 & 0.3 & 0.4 & 0.5 & 0.6 & 0.7 & 0.8 & 0.9 \\
\midrule
\multirow{9}{*}{QA}
& DeepSeek-R1-14B  & 31.1 & 33.0 & 34.8 & 36.7 & 38.6 & 40.5 & 42.4 & 44.2 & 46.1 \\
& GPT-4.1 mini     & 16.9 & \textbf{16.6} & \textbf{16.2} & \textbf{15.9} & \textbf{15.6} & \textbf{15.2} & \textbf{14.9} & \textbf{14.6} & \textbf{14.2} \\
& Qwen2.5-32B      & 62.0 & 58.2 & 54.4 & 50.6 & 46.8 & 43.1 & 39.3 & 35.5 & 31.7 \\
& Gemma-2-27B      & 72.3 & 66.1 & 59.9 & 53.7 & 47.5 & 41.2 & 35.0 & 28.8 & 22.6 \\
& Mistral-nemo-12B & 26.8 & 33.5 & 40.3 & 47.1 & 53.8 & 60.6 & 67.4 & 74.2 & 80.9 \\
& LLaMA-3.1-8B     & 74.8 & 68.1 & 61.3 & 54.5 & 47.7 & 40.9 & 34.1 & 27.4 & 20.6 \\
& TigerLLM-9B      & 36.2 & 34.2 & 32.2 & 30.3 & 28.3 & 26.3 & 24.3 & 22.4 & 20.4 \\
& TituLLM-3B       & \textbf{10.2} & 20.2 & 30.1 & 40.1 & 50.0 & 59.9 & 69.9 & 79.9 & 89.8 \\
& BanglaLLaMA-13B  & 20.9 & 27.9 & 34.8 & 41.7 & 48.6 & 55.5 & 62.5 & 69.4 & 76.3 \\
\midrule
\multirow{9}{*}{Code-Mix}
& DeepSeek-R1-14B  & 16.4 & \textbf{19.0} & 21.7 & 24.3 & 26.9 & 29.6 & 32.2 & 34.8 & 37.5 \\
& GPT-4.1 mini     & 20.2 & 20.6 & \textbf{21.1} & \textbf{21.5} & \textbf{21.9} & \textbf{22.3} & 22.7 & 23.2 & 23.6 \\
& Qwen2.5-32B      & 29.3 & 28.0 & 26.6 & 25.3 & 24.0 & 22.7 & \textbf{21.4} & \textbf{20.0} & \textbf{18.7} \\
& Gemma-2-27B      & 36.3 & 34.5 & 32.6 & 30.8 & 29.0 & 27.1 & 25.3 & 23.5 & 21.6 \\
& Mistral-nemo-12B & 31.0 & 37.1 & 43.2 & 49.3 & 55.4 & 61.5 & 67.6 & 73.7 & 79.8 \\
& LLaMA-3.1-8B     & 71.6 & 65.2 & 58.8 & 52.4 & 45.9 & 39.5 & 33.1 & 26.6 & 20.2 \\
& TigerLLM-9B      & 36.1 & 35.0 & 33.8 & 32.7 & 31.5 & 30.4 & 29.2 & 28.1 & 26.9 \\
& TituLLM-3B       & \textbf{11.7} & 21.5 & 31.2 & 40.9 & 50.6 & 60.3 & 70.1 & 79.8 & 89.5 \\
& BanglaLLaMA-13B  & 18.7 & 26.1 & 33.5 & 41.0 & 48.4 & 55.8 & 63.2 & 70.7 & 78.1 \\
\midrule
\multirow{9}{*}{Summ.}
& DeepSeek-R1-14B  & \textbf{7.8} & \textbf{8.9} & 10.0 & 11.1 & 12.2 & 13.3 & 14.3 & 15.4 & 16.5 \\
& GPT-4.1 mini     & 11.1 & 10.7 & 10.4 & 10.1 &  9.8 &  9.5 &  9.2 &  8.8 &  8.5 \\
& Qwen2.5-32B      & 10.1 &  9.5 & \textbf{8.9} & \textbf{8.3} & \textbf{7.7} & \textbf{7.1} & \textbf{6.5} & \textbf{5.9} & \textbf{5.3} \\
& Gemma-2-27B      & 28.7 & 26.0 & 23.2 & 20.5 & 17.8 & 15.1 & 12.4 &  9.6 &  6.9 \\
& Mistral-nemo-12B & 15.9 & 22.0 & 28.1 & 34.1 & 40.2 & 46.2 & 52.3 & 58.4 & 64.4 \\
& LLaMA-3.1-8B     & 26.4 & 27.0 & 27.6 & 28.2 & 28.8 & 29.4 & 30.0 & 30.6 & 31.2 \\
& TigerLLM-9B      & 43.3 & 39.0 & 34.7 & 30.4 & 26.2 & 21.9 & 17.6 & 13.3 &  9.1 \\
& TituLLM-3B       & 10.3 & 20.3 & 30.2 & 40.2 & 50.2 & 60.1 & 70.1 & 80.1 & 90.0 \\
& BanglaLLaMA-13B  & 13.1 & 21.5 & 30.0 & 38.4 & 46.8 & 55.3 & 63.7 & 72.1 & 80.6 \\
\midrule
\multirow{9}{*}{Math.\ Reas.}
& DeepSeek-R1-14B  & 16.7 & \textbf{17.0} & \textbf{17.4} & \textbf{17.7} & \textbf{18.1} & \textbf{18.5} & \textbf{18.8} & 19.2 & 19.5 \\
& GPT-4.1 mini     & 22.2 & 23.2 & 24.3 & 25.3 & 26.4 & 27.5 & 28.5 & 29.6 & 30.6 \\
& Qwen2.5-32B      & 28.4 & 28.0 & 27.6 & 27.2 & 26.8 & 26.4 & 26.0 & 25.6 & 25.2 \\
& Gemma-2-27B      & 65.9 & 59.2 & 52.5 & 45.8 & 39.0 & 32.3 & 25.6 & \textbf{18.9} & 12.2 \\
& Mistral-nemo-12B & 87.5 & 77.9 & 68.4 & 58.8 & 49.3 & 39.8 & 30.2 & 20.7 & \textbf{11.1} \\
& LLaMA-3.1-8B     & 82.0 & 74.8 & 67.7 & 60.5 & 53.3 & 46.1 & 38.9 & 31.8 & 24.6 \\
& TigerLLM-9B      & 79.1 & 72.7 & 66.3 & 59.9 & 53.5 & 47.0 & 40.6 & 34.2 & 27.8 \\
& TituLLM-3B       & \textbf{10.0} & 20.0 & 30.0 & 40.0 & 50.0 & 60.0 & 70.0 & 80.0 & 90.0 \\
& BanglaLLaMA-13B  & \textbf{10.0} & 20.0 & 30.0 & 40.0 & 50.0 & 60.0 & 70.0 & 80.0 & 90.0 \\
\bottomrule
\end{tabular}
\caption{Nine-point weight sweep of
$\text{BHS}_w = w_A(\text{A-err.}) + (1-w_A)(\text{B-err.})$ for every model and task
(zero-shot). \textbf{Bold} = best (lowest) model at that weighting; TituLLM-3B and
BanglaLLaMA-13B tie for best on Mathematical Reasoning at $w_A = 0.1$. Rankings are stable
through the central range; at the extremes the nominal winner is in almost every case a
degenerate responder.}
\label{tab:weight_sweep}
\end{table*}

\section{Additional Bengali-Centric Models and Expanded CoT}
\label{app:bengali_models}
We evaluate TituLLM-3B \cite{nahin2025titullms} (\texttt{hishab/titulm-llama-3.2-3b-v1.1})
and BanglaLLaMA-13B \cite{zehady2026banglallama}
(\texttt{BanglaLLM/bangla-llama-13b-instruct-v0.1}) on a stratified 10\% sample across
all four tasks under the identical dual-track protocol.

\section{Human Validation Study}
\label{app:human_val}
Three undergraduate annotators, all native Bengali speakers who have completed written and
spoken Bangla coursework to the intermediate level, independently annotated a stratified
sample spanning all four tasks and all twelve hallucination types, containing both
hallucinated candidates and gold instances in benchmark proportions. Annotation was blind
to hallucination-type and source labels, and ground truth was set by majority vote (2 of 3).

\begin{table}[h]
\centering\small
\begin{tabular}{lcccc}
\toprule
\textbf{Task} & \textbf{Items} & \textbf{Fleiss' $\kappa$} & \textbf{Exact} & \textbf{Unanim.} \\
\midrule
QA                  & 241/250 & 0.918 & 97.2\% & 95.9\% \\
Code-Mixed QA       & 250/250 & 0.914 & 98.4\% & 97.6\% \\
Summarization       & 200/200 & 0.926 & 97.3\% & 96.0\% \\
Math.\ Reasoning    & 300/300 & 0.911 & 97.3\% & 96.0\% \\
\bottomrule
\end{tabular}
\caption{Inter-annotator agreement. Nine QA items were excluded because one annotator
skipped them. By the Landis--Koch interpretation, $\kappa = 0.911$--$0.926$ is
``almost perfect'' agreement.}
\label{tab:iaa}
\end{table}

\paragraph{Code-mixed conversion authenticity.} The same annotators reviewed the
GPT-generated code-mixed conversions along two axes. Meaning preservation reached
Fleiss' $\kappa = 0.906$, with meaning preserved in 92.9\% of scored conversions; the
remaining items were unanimously flagged and have been revised or excluded. Naturalness,
rated on a 1--5 Likert scale, averaged 4.38, with 88.9\% of conversions rated 4 or higher.

\section{Seed Selection Analyses}
\label{app:seed_analyses}

\subsection{Cross-Family Judge Check}
\label{app:cross_judge}
\begin{table}[h]
\centering\small
\begin{tabular}{lcc}
\toprule
\textbf{Generator} & \textbf{Agreement} & \textbf{Differs} \\
\midrule
Qwen2.5-14B     & 476/500 (95.2\%) & 24 \\
DeepSeek-R1-14B & 462/500 (92.4\%) & 38 \\
Gemma-2-9B      & 456/500 (91.2\%) & 44 \\
\bottomrule
\end{tabular}
\caption{Agreement between the Qwen2.5-32B judge and an independent GPT-4.1 mini judge
on a random 500-instance subsample (seed 42), identical prompt. Disagreement is lowest
for the same-family generator, the opposite of a self-preference effect.}
\label{tab:cross_judge}
\end{table}

\subsection{BERTScore and Faithfulness Are Independent}
\label{app:nli}
Scoring the summarization pool with a multilingual NLI entailment metric (summary against
source document) yields a near-zero rank correlation with BERTScore F1 (Spearman
$\rho = -0.06$). Selecting the seed pool by NLI instead of BERTScore would change roughly
half the documents (495 of 1,000, close to the overlap expected by chance), confirming the
two criteria are effectively independent. The same NLI metric separates hallucinated from
gold summaries at AUC $= 0.90$, confirming the constructed candidates are genuine
faithfulness violations regardless of how the source documents were selected.

\subsection{Reasoning Chain Length and Detection Difficulty}
\label{app:chain_length}
Splitting the 1,000 reasoning items at the median chain length and averaging across the
evaluated judges, the Track A false-alarm rate rises from 11.3\% to 13.1\% ($p = 0.014$)
and the Track B miss rate from 56.1\% to 59.3\% ($p = 0.001$), raising BenHalluScore by
2.6 points on the longer half. Comparing the shortest quarter with the longest quarter
widens the gap (miss rate 55.5\% $\rightarrow$ 60.2\%, $p < 0.001$), and BenHalluScore
rises monotonically across ten length bins (33.4\% $\rightarrow$ 37.6\%). Because every
item already comes from the longer chains in SOMADHAN (our shortest and longest chains
differ in mean length by only $1.72\times$), these figures are a conservative lower bound.

\end{document}